\documentclass{article}

\usepackage{arxiv}
\usepackage[utf8]{inputenc} %
\usepackage[T1]{fontenc}    %
\usepackage{hyperref}       %
\usepackage{url}            %
\usepackage{booktabs}       %
\usepackage{amsfonts}       %
\usepackage{nicefrac}       %
\usepackage{microtype}      %
\usepackage{graphicx}
\usepackage{comment}
\usepackage{subcaption}
\usepackage{tabularx}
\usepackage{multirow} %
\usepackage{booktabs}
\usepackage{longtable}
\usepackage{tikz}
\usepackage[giveninits=true, backend=biber, style=numeric-comp, sorting=none, minbibnames=5, maxbibnames=7, isbn=false,url=true,eprint=false, doi=false]{biblatex}
\bibliography{references.bib}
\usepackage{svg} %

\usepackage{doi}
\usepackage{xcolor}
\usepackage[acronym]{glossaries}
\usepackage{float}
\usepackage{enumitem}

\newcommand{\overbar}[1]{\mkern 1.5mu\overline{\mkern-1.5mu#1\mkern-1.5mu}\mkern 1.5mu}

\title{Gaussian process-based online health monitoring and fault analysis of lithium-ion battery systems from field data}

\author{
\href{https://orcid.org/0000-0001-8767-4101}{\includegraphics[scale=0.06]{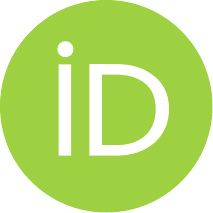}\hspace{1mm}Joachim Schaeffer}\\
Control and Cyber-Physical Systems Laboratory \\
TU Darmstadt, Germany \\
\And
\href{https://orcid.org/0000-0003-1742-4750}{\includegraphics[scale=0.06]{orcid.pdf}\hspace{1mm}Eric Lenz}\\
Control and Cyber-Physical Systems Laboratory \\
TU Darmstadt, Germany \\
\And
Duncan Gulla\\
Control and Cyber-Physical Systems Laboratory \\
TU Darmstadt, Germany \\
\And
\href{https://orcid.org/0000-0002-8200-4501}{\includegraphics[scale=0.06]{orcid.pdf}\hspace{1mm}Martin Z. Bazant}\\
Massachusetts Institute of Technology\\
Cambridge, MA, USA\\
\And
\href{https://orcid.org/0000-0003-4304-3484}{\includegraphics[scale=0.06]{orcid.pdf}\hspace{1mm}Richard D. Braatz}\\
Massachusetts Institute of Technology\\
Cambridge, MA, USA\\
\texttt{\scriptsize braatz@mit.edu}
\And
\href{https://orcid.org/0000-0002-9112-5946}{\includegraphics[scale=0.06]{orcid.pdf}\hspace{1mm}Rolf Findeisen}\\
Control and Cyber-Physical Systems Laboratory \\
TU Darmstadt, Germany \\
\texttt{\scriptsize rolf.findeisen@iat.tu-darmstadt.de} \\
}

\renewcommand{\headeright}{}
\renewcommand{\undertitle}{}
\renewcommand{\shorttitle}{This version is outdated. Final article (open access): \url{https://doi.org/10.1016/j.xcrp.2024.102258}}
\hypersetup{
pdftitle={Gaussian process-based online battery system health monitoring},
}
\newacronym{bms}{BMS}{Battery Management System}
\newacronym{ev}{EV}{Electric Vehicle}
\newacronym{soh}{SOH}{State Of Health}
\newacronym{soc}{SOC}{State Of Charge}
\newacronym{ecm}{ECM}{Equivalent Circuit Model}
\newacronym{eis}{EIS}{Electrochemical Impedance Spectroscopy}
\newacronym{ml}{ML}{Machine Learning}
\newacronym{lfp}{LFP}{Lithium-Iron-Phosphate}
\newacronym{ocv}{OCV}{Open Circuit Voltage}
\newacronym{dod}{DOD}{Depth Of Discharge}
\newacronym[longplural={Gaussian Processes}]{gp}{GP}{Gaussian Process}
\newacronym{gpr}{GPR}{Gaussian Process Regression}
\newacronym{eos}{EOS}{End of Service}
\newacronym{rls}{RLS}{Recursive Least Squares}
\newacronym{ekf}{EKF}{Extended Kalman Filter}
\newacronym{rbf}{RBF}{Radial Basis Function}
\newacronym{tp}{TP}{True Positive}
\newacronym{fp}{FP}{False Positive}
\newacronym{tn}{TN}{True Negative}
\newacronym{fn}{FN}{False Negative}
\newacronym{knn}{KNN}{K-Nearest-Neighbor}
\newacronym{eol}{EOL}{End Of Life}
\newacronym{rgp}{RGP}{Recursive Gaussian Process}
\newacronym{dt}{DT}{Decision Tree}
\newacronym{op}{OP}{Operating Point}
\newacronym[longplural={Lithium-Ion Batteries}]{lib}{LIB}{Lithium-Ion Battery}
\newacronym{si}{SI}{Supplemetary Information}
\newacronym{p2d}{P2D}{Pseudo-Two-Dimensional}
\newacronym{spm}{SPM}{Single-Particle Model}
\newacronym{se}{SE}{Squared Exponential}
\newacronym{wv}{WV}{Wiener Velocity}

\begin{document}
\maketitle
\vspace{1cm}
{\LARGE This version is outdated. The final article is published as open access in Cell Reports Physical Science.}
\vspace{0.5cm}\\
{\LARGE Article (open): \url{https://doi.org/10.1016/j.xcrp.2024.102258}}
\vspace{0.3cm}\\
{\LARGE BattGP Code: \url{https://github.com/JoachimSchaeffer/BattGP}}
\vspace{0.5cm}\\
{\LARGE Data set: \url{https://zenodo.org/records/13715694}}

\newpage
\begin{abstract}
Health monitoring, fault analysis, and detection are critical for the safe and sustainable operation of battery systems. We apply Gaussian process resistance models on lithium iron phosphate battery field data to effectively separate the time-dependent and operating point-dependent resistance. The data set contains 29 battery systems returned to the manufacturer for warranty, each with eight cells in series, totaling 232 cells and 131 million data rows.

We develop probabilistic fault detection rules using recursive spatiotemporal Gaussian processes. These processes allow the quick processing of over a million data points, enabling advanced online monitoring and furthering the understanding of battery pack failure in the field.
The analysis underlines that often, only a single cell shows abnormal behavior or a knee point, consistent with weakest-link failure for cells connected in series, amplified by local resistive heating.  The results further the understanding of how batteries degrade and fail in the field and demonstrate the potential of efficient online monitoring based on data. 

We open-source the code and publish the large data set upon completion of the review of this article.
\end{abstract}
\keywords{Lithium-Ion Batteries  \and Field Data \and Gaussian Processes \and Machine Learning \and Artificial Intelligence \and Health Monitoring \and Fault Detection}

\vspace{1cm}
\Glspl{lib} are essential for \glspl{ev}, grid storage, mobile applications, and consumer electronics. Over the last 30 years, remarkable advances have led to long-lasting cells with high energy efficiency and density \cite{blomgren2016development}. The growth of production volume over the last decade is projected to continue \cite{legal_battvo_eu_23, mcKinsey_BatteryProduction} mainly due to \glspl{ev} and stationary storage, both needed for the transition to a sustainable future. Safe operation of \glspl{lib} is vital to protect life and property and to strengthen trust in \glspl{lib}. In the past, \gls{lib} fires erupted in many different applications, including \glspl{ev} \cite{sun2020review_fires}, stationary storage \cite{social_cosntruction_koreaBESSfire2023}, and electric bicycles \cite{new_york_ebike_fires}. 
Monitoring batteries during operation allows the detection of electrical or mechanical abuse of the system or the onset of accelerated cell degradation, which is critical to reducing the potential of such fires.  

This article considers the design of \gls{gp}-based health monitoring systems from battery field data, which are time series data consisting of noisy temperature, current, and voltage measurements corresponding to the system, module, and cell level \cite{sulzer2021challenge}. 
In real-world applications, the operational conditions are usually uncontrolled, i.e., the device is in the hands of the customer, who can use and potentially abuse the battery system. From a high level, the key task for the \gls{bms} is to ensure safe operation of the battery system \cite{plett2015BMS_vol2}, either onboard or potentially also leveraging the cloud \cite{li2020digital} if further investigations or compute power are needed. Field data are critical for improving \glspl{bms}, understanding how batteries age in the field, and detecting faults early under realistic battery operating conditions \cite{sulzer2021challenge, pozzato2023analysis}. Furthermore, openly available health monitoring methods are important for regulation, such as the battery passport \cite{legal_battvo_eu_23}.
A critical bottleneck in pushing the field of battery monitoring forward has been the lack of published battery field data \cite{ward2022genome}, which is one of the contributions of this article.

Many methods are available for modeling the cycling and degradation behavior of battery cells and systems \cite{schaeffer2024ACC, krewer2018dynamic,ramadesigan2012modeling}. Roughly, these battery models can be categorized as empirical models (e.g., \cite{petit2016development, samadani2015empirical}), machine learning models (e.g., \cite{severson2019data, jones_penelope_2021_5704796, schaeffer_ml_benchmarks_EIS_QS, schaeffer2024interpretationNullspace, RICHARDSON2017209, richardson_2019_cap_estimation}), first-principles models (aka physics-based aka mechanistic, e.g., \cite{newman1975porous, doyle1993modeling, jokar2016review}), and hybrid models which combine first-principles and machine learning models (e.g., \cite{aykol2021perspective, tu2023integrating, NASCIMENTO2021230526}). Choosing an appropriate model is essential to use the available data optimally \cite{schaeffer2024ACC}. 
Physics-based battery models such as \gls{p2d} and \glspl{spm} are often challenging to parameterize with field data where often no or only a few cell parameters are known in combination with sensor noise and bias as well as low time resolution. \Glspl{ecm} are an alternative because they are easier to parameterize with limited data \cite{krewer2018dynamic} and are applicable to field data \cite{aitioSolar2021}. 

Faults are abnormal events that cause the system to behave in an unintended way or stop operating. Furthermore, faults can potentially cause safety threats to a system and its environment. Fault detection methods can be categorized as signal-based or model-based. Much research considers the fast signal-based fault detection for battery systems \cite{hu2020advancedfault_batteries_review, wang2019review_failure_fire_prevention, tran2020review}. A few examples of commonly used methods include normalized voltage-based methods \cite{data_driven21_norm_voltage}, analysis of correlation coefficients of cell voltages \cite{kang2020online, li2018novel_icc}, and sample entropy-based methods \cite{li2020lithium_sample_entropy}. Model-based fault detection methods are more computationally complex but can potentially detect faults earlier and improve robustness (see \cite{Chen_Xiong_Tian_Shang_Lu_2016, zhang2021multi, kalman_filter_modeling_fault, zhao2017fault, samanta2021ml_fault_review_lion, schmid_data_driven_2021, zhao2022data}). Battery system faults can be auxiliary, sensor, or battery faults. Battery faults are often due to slow processes, such as battery degradation, which can be addressed by real-time estimation of the \gls{soh} of batteries to enable prediction of battery lifetime \cite{aitioSolar2021, xiong2018towards, berecibar2016critical}. 
It is commonly observed that slow capacity decay is followed by accelerated degradation, yielding a knee-shaped capacity decay curve as a function of charge throughput \cite{attia2022knees}. 

In this work, we analyze and model lithium-ion battery systems based on field data using a hybrid approach of machine learning and \glspl{ecm}. 
Inspired by \cite{aitioSolar2021}, we develop a \gls{gp}-based resistance modeling framework for lithium-ion battery systems without the need for an \gls{ocv} curve for \gls{lfp} batteries. We showcase exact \gls{gp} results using 40k data points for each cell and recursive spatiotemporal \gls{gp} results based on all selected data points (above 1 million data points for the system with the most available data). Furthermore, we develop fault probabilities, allowing probabilistic monitoring of the time-dependent internal resistance and associated fault probabilities for each individual cell and the whole system.
The derivations are motivated by field data from 29 portable lithium-ion battery systems consisting of eight cells in series, totaling 232 cells that were in use for up to 5 years. All analyzed systems were returned to the manufacturer for warranty claims, where the data were recovered. An in-depth analysis of the data set and modeling results allows us to further the understanding of how battery packs fail. This article makes the field data and associated Python package \textbf{\textit{BattGP}} available as open source. This data set provides a unique opportunity to develop and test approaches for estimating \gls{soh} and fault detection and health monitoring.
To the best of the author's knowledge, this is the first large publicly available lithium-ion battery field data set containing data from many independent systems and multiple years of operation.

\section*{Field Data Set}
\label{sec:data}
The data set contains data from 29 portable 24\,V \gls{lfp} battery systems with approximately 160\,Ah nominal capacity. Each system's specific use case is unknown, but battery systems of this size are typically used as power sources for recreational vehicles, solar energy storage, and more. %

All battery systems in this data set showed some form of unsatisfactory behavior and were returned to the manufacturer. Many reasons can cause a consumer to return a battery to the manufacturer for maintenance. The user's individual decisions may be motivated by personal judgment, \gls{bms} warnings, or customer support advice. This data set comprises a very small fraction of batteries sold of this version. Therefore, this data set is biased and not representative of the operational data of the entire population of this system version. This article's battery system type was replaced by an improved version. The battery system manufacturer provided the data set for this study and allowed its open-source release under the condition of anonymity.

Some time series data contain data gaps, either because the system was fully switched off, the user tampered with it and its data storage unit, or for other unknown reasons. Furthermore, the exact manufacturing and shipping dates are unavailable, making it impossible to reconstruct charge throughput and rest periods' duration precisely. 
\begin{figure}[htb]
    \centering
    \includegraphics[width=0.60\textwidth]{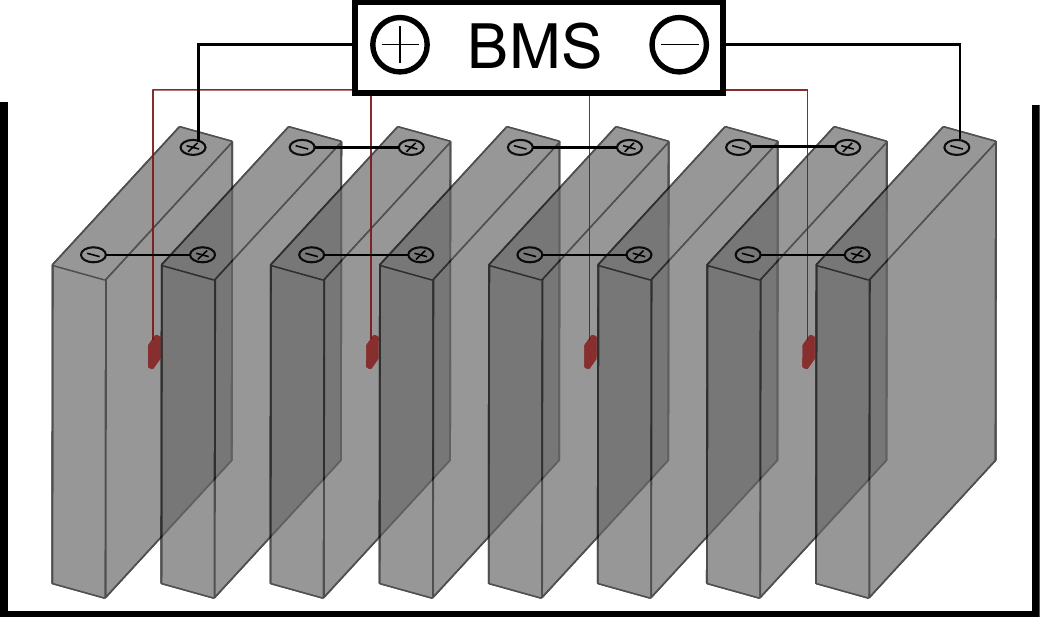}
    \caption{Sketch of a battery system with 8 prismatic cells temperature sensors shared by two cells in red.}
    \label{fig:battery_pack}
\end{figure}

\subsection*{Battery System}
\begin{figure}[hbt] 
  \begin{subfigure}[b]{0.45\linewidth}
    \centering
    \begin{tikzpicture}
    \node[anchor=south west] at (0,0) (image1) {\includegraphics[width=0.88\linewidth]{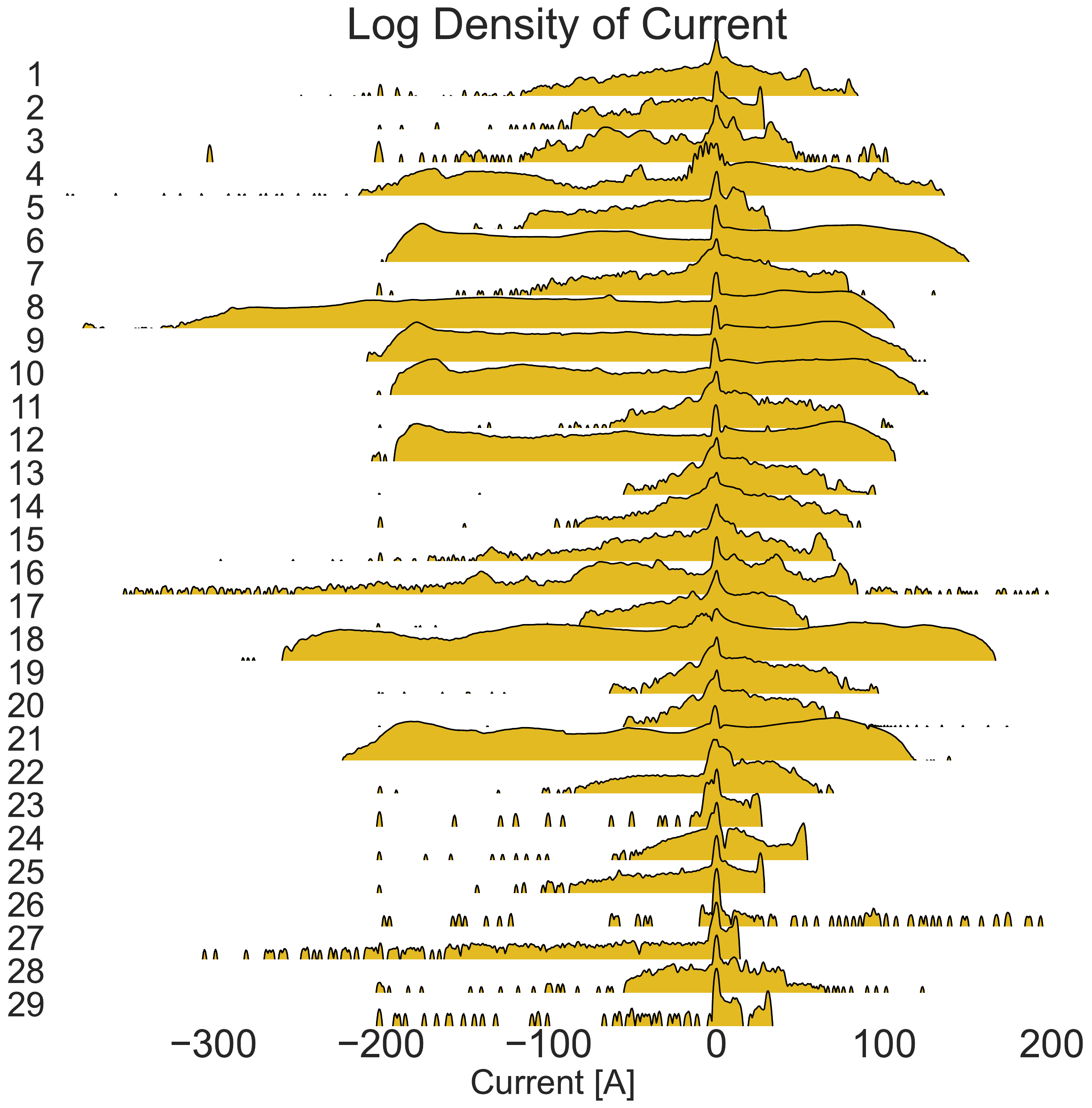}};
        \begin{scope}[x={(image1.south east)},y={(image1.north west)}]
        \node[fill=none] at (0,1) {\textbf{a)}};
        \end{scope}
    \end{tikzpicture} 
    \label{fig7:a} 
    \vspace{4ex}
  \end{subfigure}%
  \begin{subfigure}[b]{0.45\linewidth}
    \centering
    \begin{tikzpicture}
    \node[anchor=south west] at (0,0) (image1) {\includegraphics[width=0.88\linewidth]{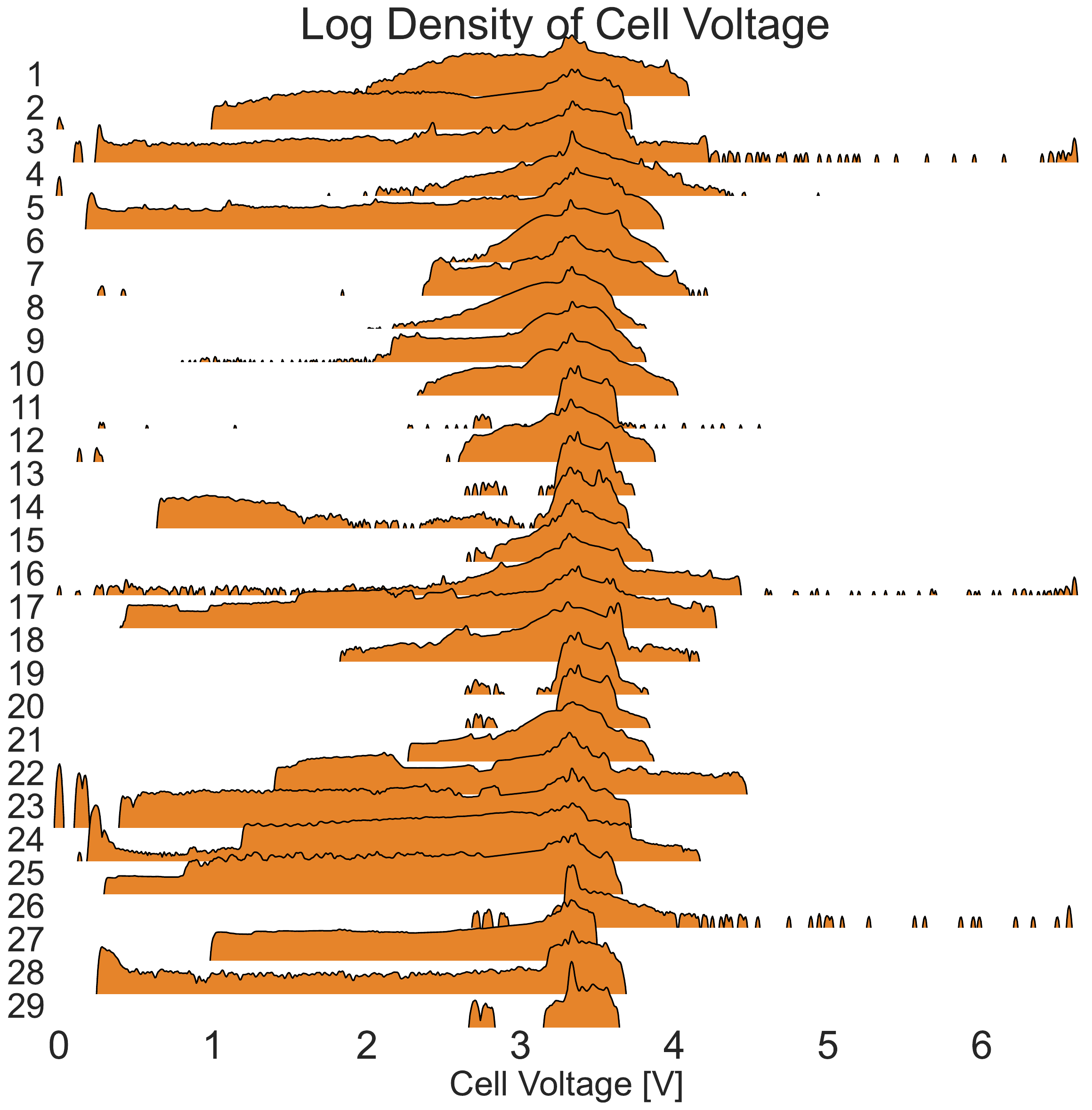}};
        \begin{scope}[x={(image1.south east)},y={(image1.north west)}]
        \node[fill=none] at (0,1) {\textbf{b)}};
        \end{scope}
    \end{tikzpicture} 
    \label{fig7:b} 
    \vspace{4ex}
  \end{subfigure} 
  \begin{subfigure}[b]{0.45\linewidth}
    \centering
    \begin{tikzpicture}
    \node[anchor=south west] at (0,0) (image1) {\includegraphics[width=0.88\linewidth]{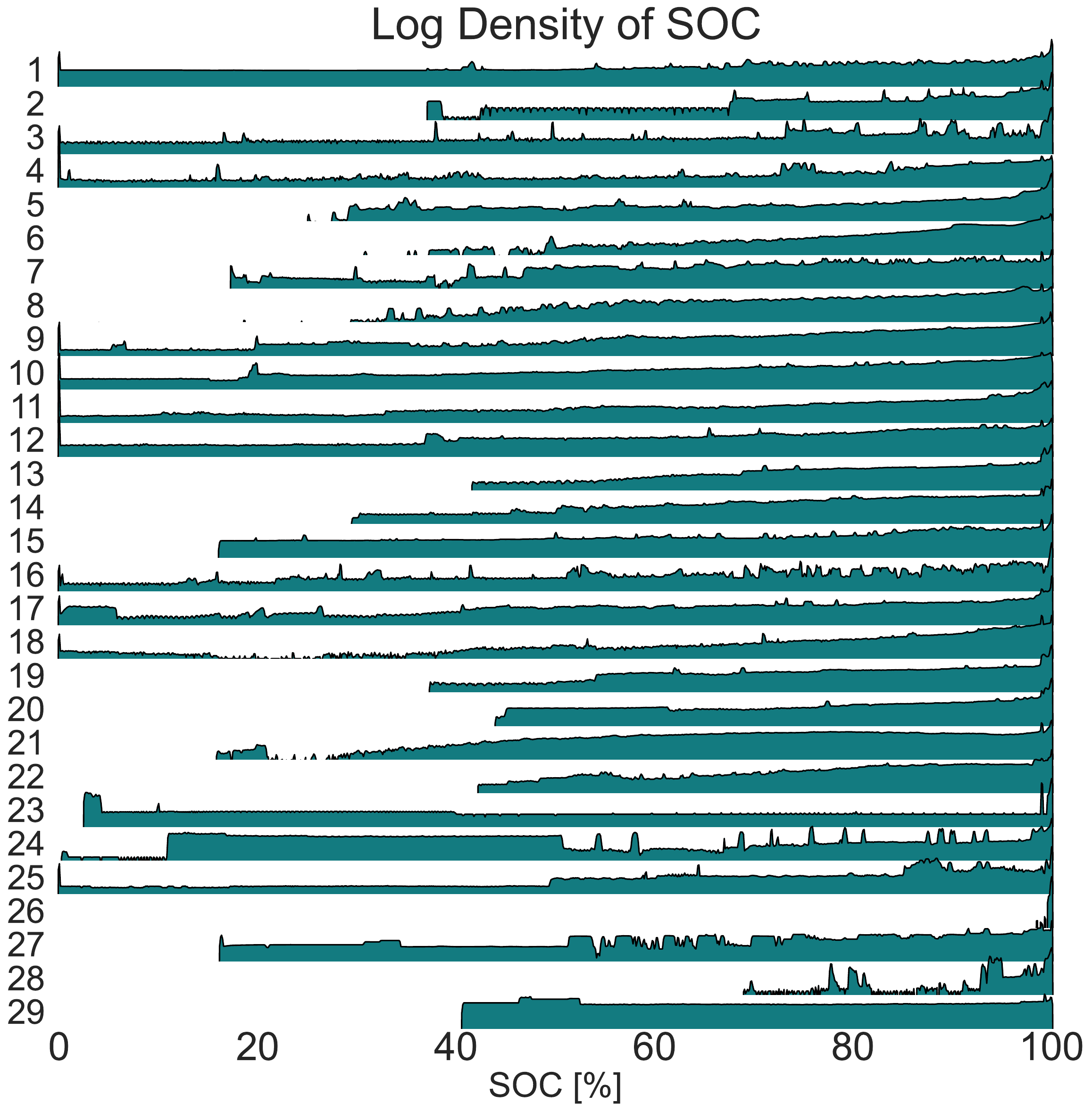}};
        \begin{scope}[x={(image1.south east)},y={(image1.north west)}]
        \node[fill=none] at (0,1) {\textbf{c)}};
        \end{scope}
    \end{tikzpicture} 
    \label{fig7:c} 
  \end{subfigure}%
  \begin{subfigure}[b]{0.45\linewidth}
    \centering
    \begin{tikzpicture}
    \node[anchor=south west] at (0,0) (image1) {\includegraphics[width=0.88\linewidth]{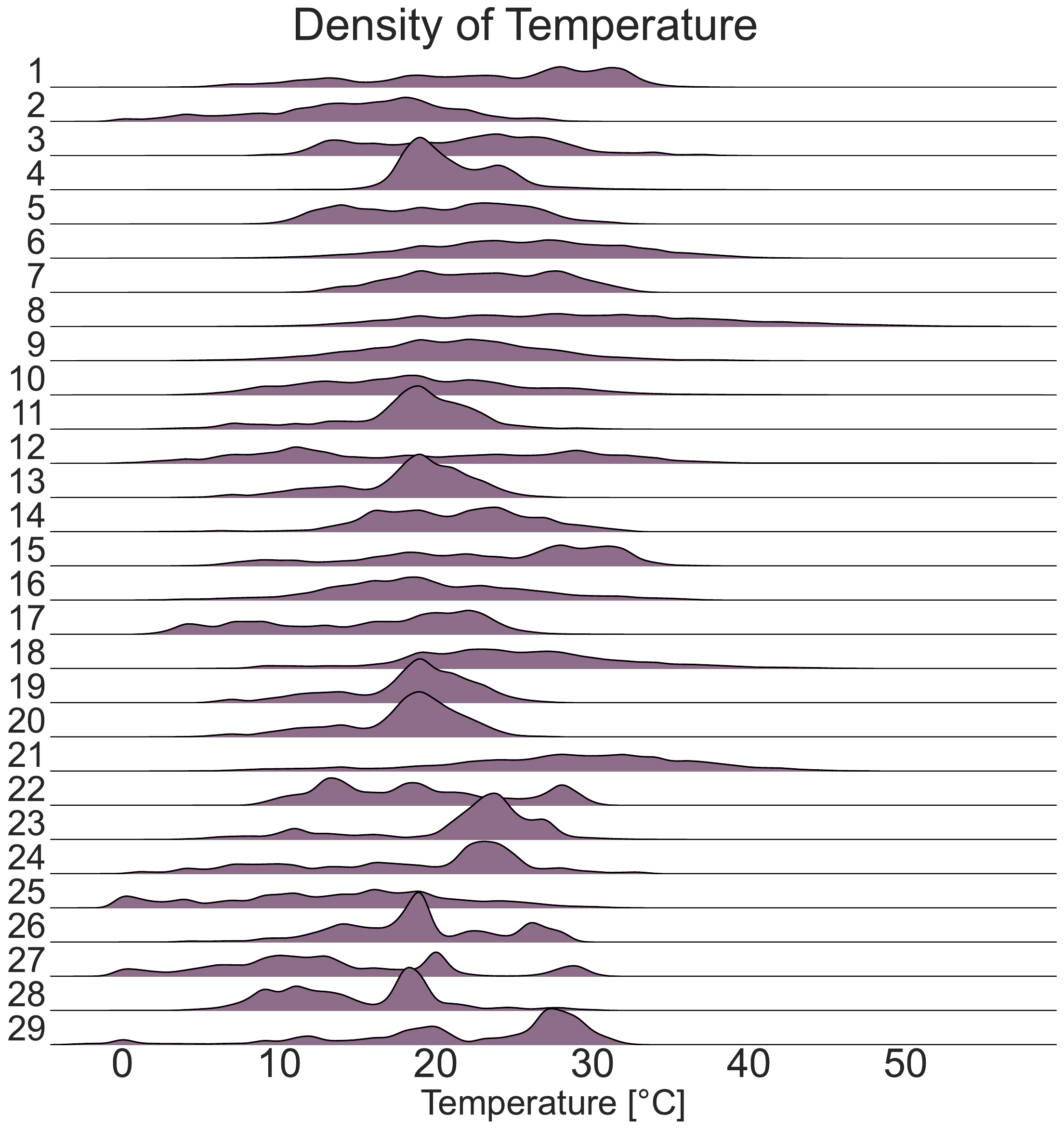}};
        \begin{scope}[x={(image1.south east)},y={(image1.north west)}]
        \node[fill=none] at (0,1) {\textbf{d)}};
        \end{scope}
    \end{tikzpicture} 
    \label{fig7:d} 
  \end{subfigure} 
  \caption{Estimations of the logarithmic probability densities of the distributions of a) current, b) cell voltages, c) \gls{soc}, and d) probability density of temperatures.}
  \label{fig:joy_us} 
\end{figure}
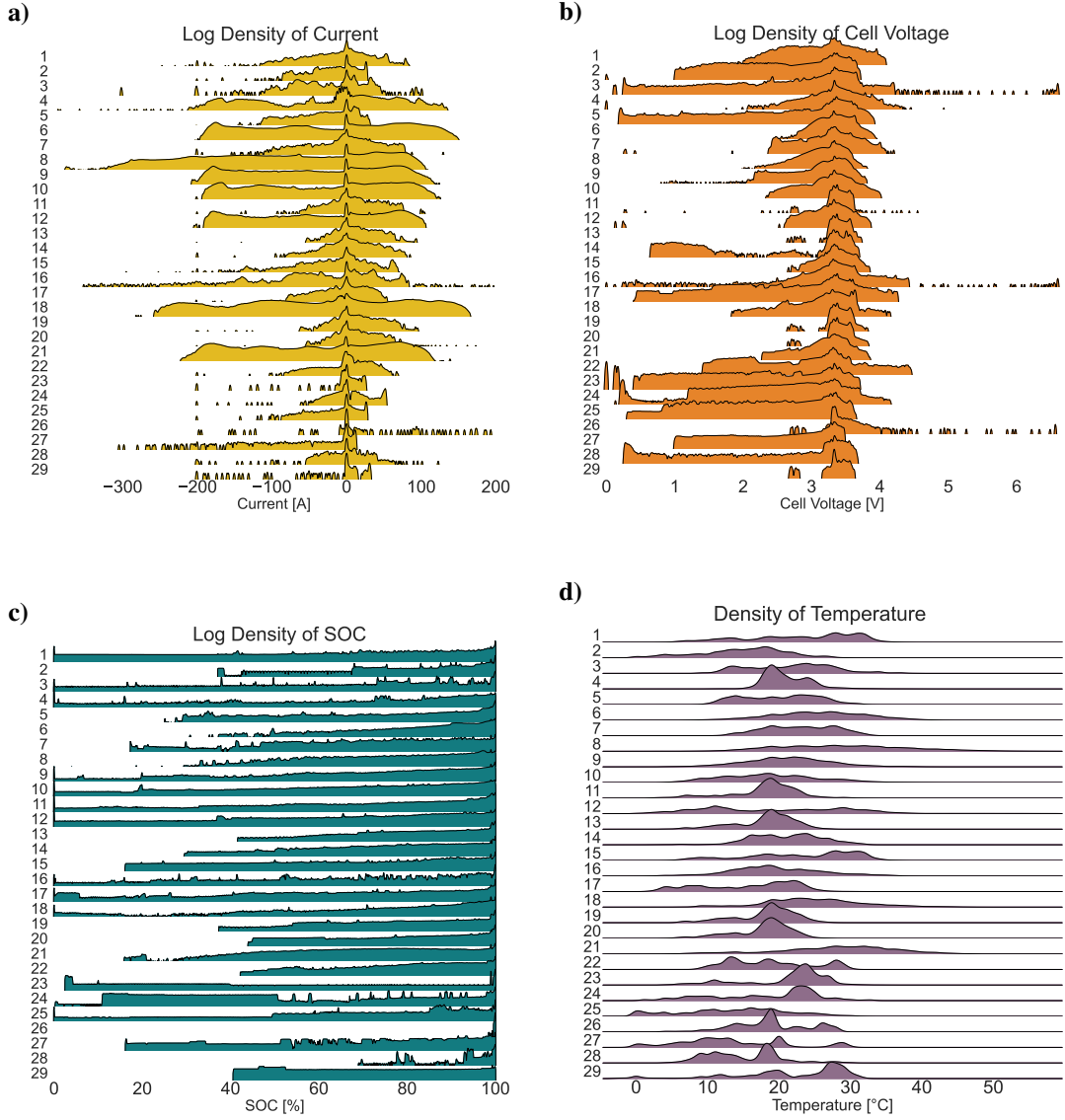

Each battery system consists of 8 prismatic cells in series. Each system has one load current sensor, and each cell has one voltage sensor. The four temperature sensors are placed between adjacent cells, i.e., each temperature sensor is shared by two cells (Fig.\ \ref{fig:battery_pack}). Furthermore, the battery systems have active cell balancing. The available measurements for the systems vary from a single month to five years. Consequently, the number of data rows per system varies from several thousand to millions, depending on the duration of battery operation. The data set contains a total of 131 million rows of measurements (cf. Tab.\,~\ref{tab:batteries}). 
\begin{table}[htb]
    \centering
    \caption{Technical specification of the battery system and summary of the associated data.}
    \label{tab:batteries}
    \vspace{1mm}
    \begin{tabular}{l|c||l|c}
    \multicolumn{2}{c}{Technical Specification} & \multicolumn{2}{c}{Data Set} \\ 
    \hline \hline
    Nominal voltage  & $24\,\mathrm{V}$         & Number of systems     & 29    \\ \hline
    Nominal capacity & $\approx 160\,\mathrm{Ah}$       & Total number of cells & 232    \\ \hline
    \# Cells  & 8 (series)               & Total rows of data    &   131\,M     \\ \hline     
    \# Current Sensors  & 1              & Median of measurement intervals    & 5\,s      \\ \hline     
    \# Voltage Sensors  & 9                        &     &       \\ \hline     
    \# Temperature Sensors & 4                     & & \\ \hline
    \# Cell Balancing Current Sensors & 8
    \end{tabular}
\end{table}
The estimated probability densities of usage conditions (Fig.\,\ref{fig:joy_us}) demonstrate the widely varying operational conditions. Note that the estimated current, voltage, and \gls{soc} densities are plotted on a logarithmic scale. For example, many temperature measurements are below room temperature, suggesting that some systems were operated outdoors or in unheated environments. %
The observed \glspl{dod} are predominantly limited to the upper half of the \gls{soc} range, and much time was spent fully charged for most systems. The distribution of voltage shows that some batteries were overcharged ($>3.6$\,V) or undercharged ($<2.0$\,V), or both at some point. Furthermore, battery systems 3, 4, and 16 have discharge current outliers $>1000$\,A, suggesting a shortcut or inrush current due to improper installation or abuse or a sensor error. It is not possible to further disentangle the reasons for these abnormally high discharge currents and over/undercharging. There is a lot of uncertainty associated with how the user installed the system and which equipment was connected because the systems are portable and suitable for many different applications.

\subsection*{Exemplary Data Visualization and Analysis}%
\begin{figure}[!tb]
\centering
    \includegraphics[trim={0.5cm 0 0.5cm 1.15cm}, clip, width=\textwidth]{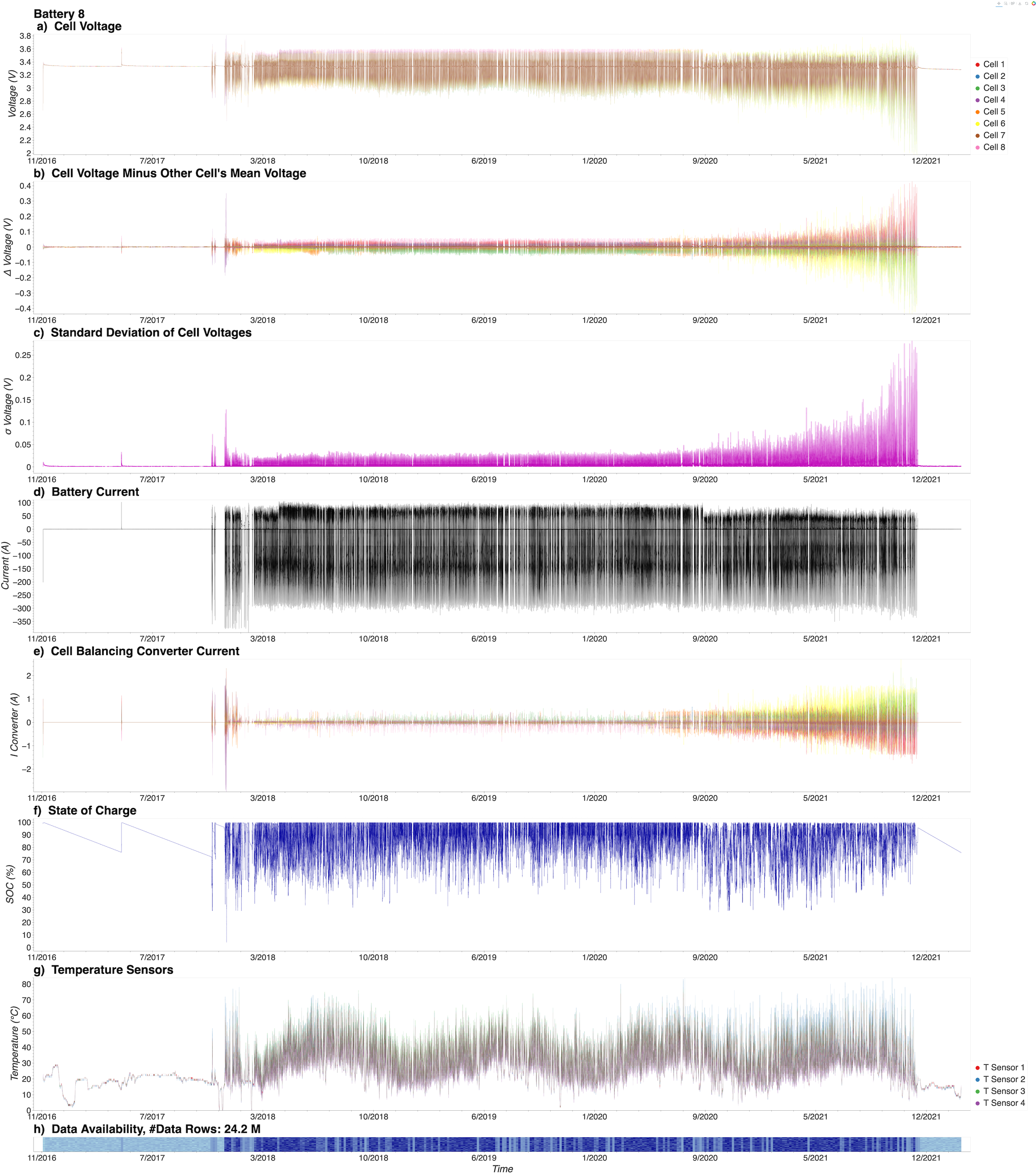}
    \caption{Data visualization of battery system \#8. a) Cell voltages, b) Cell voltages with mean of other cells subtracted, c) Cell voltage standard deviation, d) Battery system current, e) Cell balancing converter current, f) State of charge, g) Temperatures (each temperature sensor neighbors two cells (Fig.\,\ref{fig:battery_pack})), h) Data availability.}
    \label{fig:batt_id8_overview}
    \vspace{-1ex}
\end{figure}

We analyze 24.2 million rows of data associated with battery system 8, which was operated for approximately five years (Fig.\,\ref{fig:batt_id8_overview}). The system was switched on in November 2016, but frequent usage only started about a year later. The end of continuous usage is around December 2021. However, the system kept logging data for a couple more months afterward. The temperature profile shows seasonal variations with higher temperatures during the northern hemisphere summer months. Furthermore, voltage measurements and estimated \gls{soc} show that the system was primarily operated between 60 to $100\%$ \gls{soc} with occasional discharges below $40\%$. Around September 2020, the usage pattern changes, as can be seen by the current and \gls{soc} patterns. %
The mean subtracted cell voltages,
\begin{equation}
    \tilde u_i(t) = u_i(t) - \frac{1}{n-1}\sum_{j, \,j \neq i}^{n} u_j(t),
\end{equation}
where $u_i(t)$ is the voltage of cell $i$ at time $t$, and $n$ is the number of cells, show average deviations below 0.1\,V for the first two years of operation (Fig.\,\ref{fig:batt_id8_overview}b). With increasing usage, i.e., increasing charge throughput and time and therefore also degradation, the average mean subtracted voltages increase, an indicator that individual cells age differently, likely due to cell-to-cell variations (e.g., \cite{en14113276, BARBERS2024110851}), and as a consequence, the system is less balanced. Similarly, the voltage standard deviations,
\begin{equation}
    \sigma_u(t) = \sqrt{\frac{1}{n-1} \sum_{i=1}^n (u_i(t) - \mu(t))^2},\text{ where } \mu(t) = \frac{1}{n} \sum_{i=1}^n u_i(t),
\end{equation}
are below 0.05\,V for the first years; however, toward the end of use, the standard deviation of cell voltages increases significantly.

To summarize, the data analysis of system 8 shows heavy usage over a five-year period, totaling about 1531 equivalent full cycles. However, based on the data visualization, it is challenging to understand further how cells degraded, whether certain cells degraded more than others, or when the system might fail. In the next section, we explore how to use the battery data for internal state estimation suitable for tracking the aging behavior of the individual cells.  

\clearpage

\section*{Gaussian Process Equivalent Circuit Battery Systems Modeling}
\label{sec:headings}
Gaussian Processes (GPs) are suitable for modeling the time- and operating point behavior of batteries \cite{aitioSolar2021}. \Glspl{gp}, $f(x) \sim \mathrm{GP}(\mu(x), k(x, x'))$, are nonparametric probabilistic models defining a distribution of functions. \Glspl{gp} are defined by a mean function $\mu(x)$ and a covariance function $k(x, x')$ where $x, x' \in \mathcal{X}^D$ and $\mathcal{X}^D$ is the input space with dimension $D$. The modeled response $y$ follows a Gaussian distribution, as does any marginal distribution (e.g., \cite{williams2006gaussian} for further information).
\Glspl{gp} are a flexible modeling framework excelling in the case of limited data by making a point estimate and modeling the covariance associated with the prediction.
\begin{figure}[htb!]
    \centering
    \includegraphics[width=0.6\textwidth]{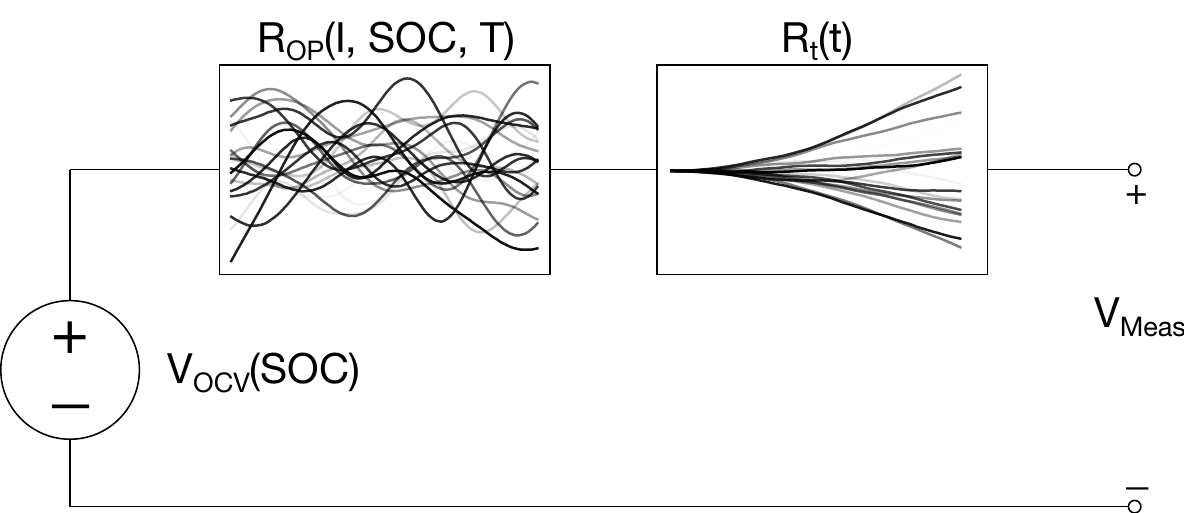}
    \caption{Visualization of the \gls{ecm} model with illustrations of random draws from the kernel functions (Radial basis function kernel with current, \gls{soc}, and temperature inputs and Wiener velocity kernel with time input, as suggested in \cite{aitioSolar2021}).}
    \label{fig:gp-ecm-model}
\end{figure}

Motivated by \cite{aitioSolar2021}, we base our analysis on the \gls{gp}-\gls{ecm} modeling framework, which uses an \gls{ecm} consisting of two series resistors modeled by two additive kernels (Fig.\ \ref{fig:gp-ecm-model}). The first resistor is modeled by an \gls{rbf} kernel taking the current, \gls{soc}, and temperature as inputs and is, therefore, only dependent on the operating point of the battery. A non-stationary  Wiener velocity kernel models the second series resistor and depends only on time. The underlying assumption of this framework is that the operating point-dependent equivalent circuit internal resistance does not depend on degradation. This assumption, while not exactly fulfilled by \glspl{lib}, is, however, reasonable, allowing us to approximate and extract the equivalent circuit time-dependent resistance, i.e., a proxy of degradation.

\paragraph{\gls{gp}-\gls{ecm} Modeling Pipeline}
The \gls{gp}-\gls{ecm} algorithm consists of the following steps for a single cell. First, voltage, current, \gls{soc}, and time are extracted and downselected. The aim is to select data points with significant information for the learning tasks and to avoid sparse extreme conditions where the dependency between the equivalent circuit operating point-dependent resistance might be more nonlinear. We use discharging sections only because, in the data set, there are more dynamics during discharging. Further selection criteria outlined in the Methods section are applied to limit the considered \gls{soc} window, discharge current range, and temperature conditions (Tab.\,\ref{tab:data_selection}). We select the latest section with no gaps larger than 100 days and require at least 600 valid data points, resulting in a selection of 21 systems for modeling. We use a linear pseudo \gls{ocv} curve roughly approximating the \gls{ocv} of \gls{lfp} to calculate the equivalent circuit resistance based on the \gls{soc} estimation of the \gls{bms} for all time steps (SI Fig.\,A.1). %
Physically, the linear \gls{ocv} model approximates the distribution of particle-size dependent nucleation barriers~\cite{cogswell2013theory}
for a porous electrode with a particular distribution of \gls{lfp} particle sizes~\cite{ferguson2014phase}.
The results indicate that the algorithm can be used even when the \gls{ocv} curve associated with the cells in the data set is not available by using a rough approximation from published \gls{ocv} curves.

Based on the pack layout, 8 cells in series, and 1 parallel string (8s1p), we generate a model for each cell, resulting in eight models for each battery system. 
We use exact \glspl{gp} to get an overview of the data set but move to recursive spatiotemporal \glspl{gp} in the next section, enabling online monitoring (see Methods section for more details).
For the exact \gls{gp}, to keep the computational complexity feasible, we further downselect each cell to a maximum of 40k measurements by selecting data with linearly spaced indices of the preselected data. The hyperparameters are optimized as outlined in the Methods section. Identical hyperparameters are used for all battery systems because the hyperparameters characterize the system behavior, and all systems are identical in construction. Even though the batteries were operated in different applications with different average operating points, \glspl{gp} can handle these differences.
\begin{figure}[H]
    \centering
    \includegraphics[width=0.99\textwidth]{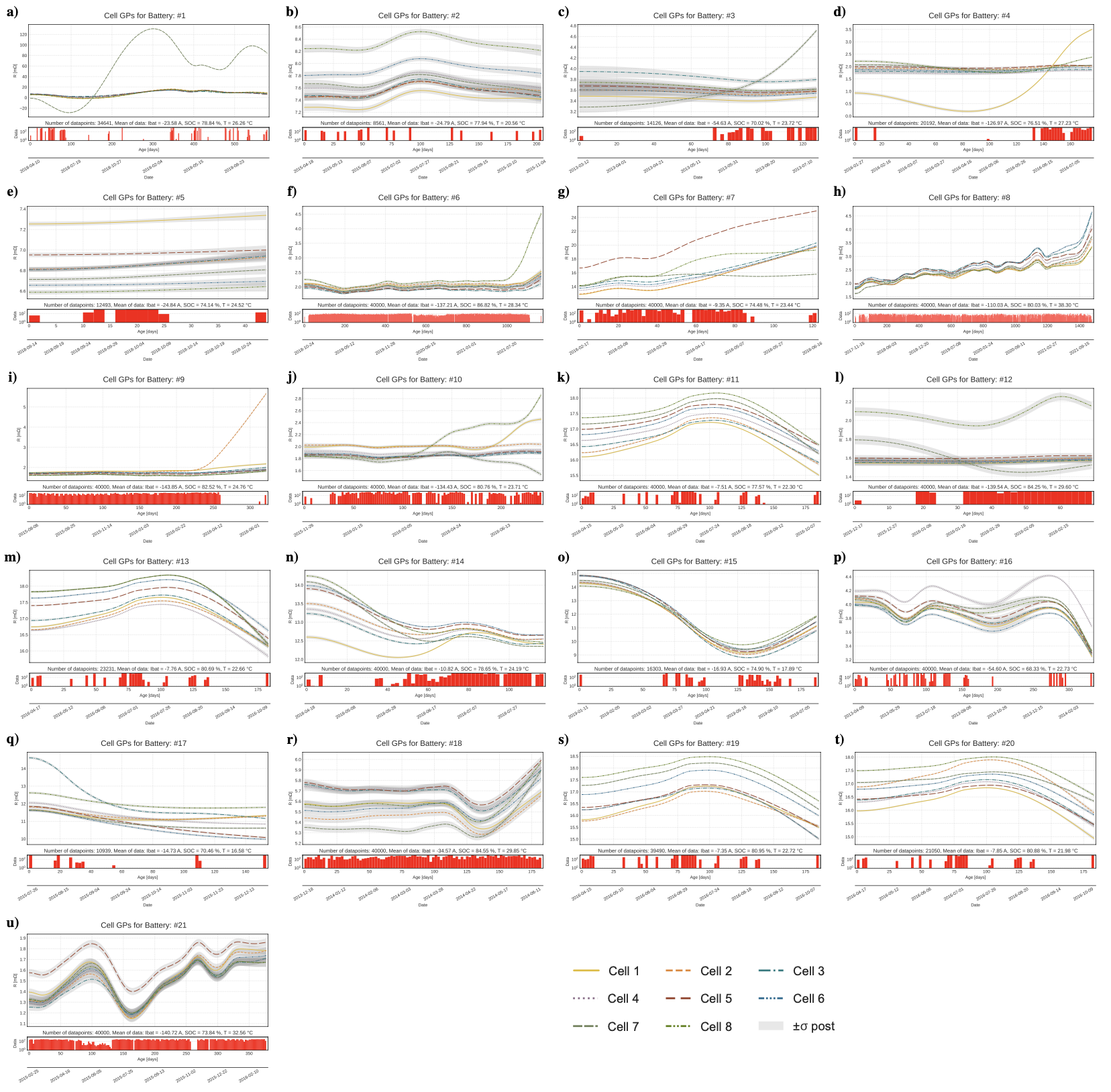}
   \caption{Evaluation of the time-dependent Wiener velocity kernel equivalent circuit resistance for up to 40k randomly selected data points after downselection.  Only systems for which >600 data points after downselection are available are shown (21 systems). Each subplot shows the equivalent circuit resistance for each cell with the reference operating point set to the mean of all selected data points (top) and a logarithmically scaled bar plot in red, visualizing the available data points over time (bottom). The plots and systems are described in more detail in Tab. \ref{tab:batteries_interpret}.}
    \label{fig:all_batts_analysis}
\end{figure}
\clearpage

\newcolumntype{b}{X}
\newcolumntype{s}{>{\hsize=.18\hsize \centering\arraybackslash}X}
\begin{table}[H]
    \centering
    \caption{Overview of \gls{gp} Analysis Results}
    \label{tab:batteries_interpret}
    \vspace{1mm}
    \begin{tabularx}{\textwidth}{ s | s | s | s | s | b }
    Subplot, System ID & Equivalent Full Cycles & Max Age (Days) & GP Section (Days) & Intriguing Cells & Comments\\
    \hline \hline
    a), 1 & 14 & 1402 & 580 &    (7)       & Cell 7 behaves differently to the other cells.  \\ \hline
    b), 2 & 6 & 1651 & 204 &     (8)       & Cell 8 shows a slightly higher resistance than the other cells. \\ \hline
    c), 3 & 11 & 591 & 127 &     (7)       & Cell 7 shows a strong upward resistance trend, while other cells have fairly constant resistances.    \\ \hline
    d), 4 & 185 & 544 & 176 &    (1)       & Cell 1 behaves very differently to other cells.  \\ \hline
    e), 5 & 28 & 2166 & 44 &   (1)         & Cell 1 shows a slightly higher resistance than the other cells.  \\ \hline
    f), 6 & 1446 & 1352 & 1170 & (8)       & Cell 8 shows a resistance kneepoint after 3 years.  \\ \hline
    g), 7 & 15 & 428 & 122 &     (5), (7), (8)       & Cell 5 has a slightly higher resistance; cells 7 and 8 show different trends after 40 days in comparison to the other cells.   \\ \hline
    h), 8 & 1531 & 1923 & 1476 &           & All cells show a knee point behavior toward the end of the section (seasonal temperature variations affecting resistance estimates, see SI\,Fig. D.8).  \\ \hline
    i), 9 & 489 & 534 & 322 & (2), (1)     & Cell 2 has a knee point after 240 days. Cell 1 has a slightly higher resistance to other cells except for cell 2.  \\ \hline
    j), 10 & 211 & 417 & 241 & (8), (1), (2), (7) & Cell 8 shows an increasing resistance after 100 days. Cells 1, 2, and 7 behave slightly differently from the other cells. \\ \hline
    k), 11 & 15 & 997 & 185 &              & Lower average current than most other systems, unclear interpretation. \\ \hline
    l), 12 & 93 & 320 & 69 &     (8), (7)       & Cells 8 and 7 show different behavior (significantly higher temperatures measured by temperature sensor 4, see SI Fig.\,D.12). \\ \hline
    m), 13 & 23 & 724 & 183 &              & Lower average current than most other systems, unclear interpretation.  \\ \hline
    n), 14 & 23 & 376 & 117 &              & Lower average current than most other systems, unclear interpretation.  \\ \hline
    o), 15 & 17 & 1217 & 187 &             & Many data gaps, unclear interpretation.  \\ \hline
    p), 16 & 34 & 799 & 332 &    (4)       & Cell 4 behaves slightly different to the other cells.  \\ \hline
    q), 17 & 21 & 889 & 153 &    (3)       & Cell 3 shows different behavior, unclear interpretation. \\ \hline
    r), 18 & 453 & 900 & 182 &             &  All cells show increasing resistance toward the end of use, unclear interpretation.\\ \hline
    s), 19 & 28 & 712 & 185 &              &  Lower average current than most other systems, unclear interpretation.\\ \hline
    t), 20 & 22 & 724 & 183 &              &  Lower average current than most other systems, unclear interpretation.\\ \hline
    u), 21 & 1088 & 644 & 377 &    (5)     &  Cell 5 behaves differently to the other cells. \\ \hline
    \end{tabularx}
\end{table}

\subsection*{Results}
\label{sec:results}
The \gls{gp} modeling results (Fig. \ref{fig:all_batts_analysis}) show the time-dependent resistance modeled by the Wiener velocity kernel at the reference operating point. The reference operating point is set to the mean of current, \gls{soc}, and temperature of the selected data for each system to keep the same reference operating point for all cells of a system. Consequently, the absolute values for the resistance vary between the systems. %
For nine systems (1, 3, 4, 6, 7, 9, 12, 16, 17), single cells or two cells behave differently than the remaining cells (Fig.\,\ref{fig:all_batts_analysis}, Tab.\,\ref{tab:batteries_interpret}). A single cell shows a knee-shaped behavior for three systems (3, 6, 9). Battery system 8 shows a knee-shaped resistance trajectory for all cells at roughly the same time. Table \ref{tab:batteries_interpret} gives an overview of equivalent full cycle count, length of available data, length of the section modeled by the \gls{gp}, intriguing cells, and comments on the resistance pattern.

There can be a wide variety of root causes for the resistance patterns. To further understand the cause of individual degradation, a mechanical inspection of the systems would be needed, which can not be done as the returned battery systems are not physically available to the authors.
Furthermore, the model cannot distinguish between battery degradation and external degradation or faults such as connector loss, corrosion, connector resistance increase, etc. If little data is present, learning the battery operating point-dependent equivalent resistance can be challenging, resulting in larger uncertainty bands. Furthermore, the output scale parameter of the Wiener velocity kernel was tuned to track battery degradation (c.f., Methods section). Therefore, it can be challenging for the \gls{gp} to capture sudden faults such as connector loss or others quickly (see SI Sec.\,B.2 for more details). %
In case of strong temperature variations, the Wiener velocity kernel can pick up some of the temperature variations, leading to downward trends (e.g., System 8). %

Systems 6, 8, and 9 have the highest equivalent full cycle count (1446, 1531, and 489). Their resistance trajectories suggest that these systems were healthy for most of their operation, with degradation accelerating towards the end of use. Judging from the presumably healthy section of these systems, the patterns of many of the other systems appear visually different, suggesting that, indeed, there was an issue during operation, consistent with the fact that the systems were returned to the manufacturer. This gives further confidence that the \gls{gp} models systems well when they are ``healthy'' and that different faults and issues show up in the modeled equivalent circuit resistance and thus show the promise of \glspl{gp} for early fault detection which we explore in the next section.%

\section*{Fault Probabilites} 
Based on the observations on the entire data set, we investigate how the four systems with the highest equivalent full cycle count, 6, 8, 9, and 10, degraded. To do so, we propose to define a battery pack consisting of cells in series to have an acceptable resistance distribution if the internal resistances of all cells are within a resistance range centered around the mean of cell resistances,
\begin{align}
    \overbar{R}_{\textrm{p},i}(t) &= \frac{1}{n-1}\sum_{j, \,j \neq i}^{n} R_j(t), \\
    p(F_{R_i}(t)) &= p(R_i(t) > \overbar{R}_{\textrm{p},i}(t) + b) + p(R_i(t) < \overbar{R}_{\textrm{p},i}(t)-b),
\end{align} %
where $\overbar{R}_{\textrm{p},i}(t)$ is the mean resistance of the time-dependent \gls{gp} resistance mean of all cells in the pack except cell $i$. $p(F_{R_i}(t))$ is the probability of a resistance fault of cell $i$, i.e., the probability of cell $i$ not being within the band $b$ of acceptable resistances.
The \gls{gp} provides the mean and probabilities needed for calculating $p(F_{R_i}(t))$. 

The \gls{gp} modeling results (Fig.\,\ref{fig:all_batts_analysis}) show that a single cell degrading strongly can define the end of life of a battery system. Therefore, we propose to define the probability of a pack resistance fault, $p(F_{R_\textrm{p}})$, as
 \begin{align}
    p(F_{R_\textrm{p}}(t)) = 1- \prod_i \left(1-p(F_{R_i}(t)(i))\right) \sim \sum_i p(F_{R_i}(t)) 
\end{align}
according to the true ``weakest link'' failure statistics for cells connected in series~\cite{gumbel1958statistics,fisher1928limiting,le2009lifetime}. In the limit of rare, independent failures dominated by a single cell, the pack resistance fault can be approximated by
\begin{equation}
    p(F_{R_\textrm{p}}(t)) \approx \max_i p(F_{R_i}(t)).
\end{equation}
The choice of $b$ depends on the acceptable cell-to-cell variations, the system design, and the reference operating point.

In an online \gls{bms} setting, the \gls{ecm}-\gls{gp} needs to be updated continuously with new data arriving. The computational complexity of exact \glspl{gp} scale with $\mathrm{O}(n^3)$, where $n$ is the number of data points, making it computationally infeasible to update the model continuously, using all selected data. Therefore, we use a recursive spatiotemporal \gls{gp}, which scales linearly with the number of data points \cite{sarkka_spatiotemp, huber2014recursive, aitioSolar2021}, allowing to process above 1 million data points within two minutes on a laptop computer, enabling the use in embedded \gls{bms} systems, or efficient calculations in the cloud for thousands of systems. Our approach is motivated by  \cite{aitioSolar2021} but not identical (see Methods section for further details). The spatiotemporal \gls{gp} walks forward in time using a Kalman filter. We use the estimate of the Kalman filter at time $t_k$, which only depends on data up to time $t_k$ to calculate forward fault probabilities.
We set $b=0.33\,\text{m}\Omega$ based on the resistance spread observed at the beginning of the life of the batteries for the systems investigated here. Furthermore, we use the mean of the mean operating points of the four systems as common reference operating point. 

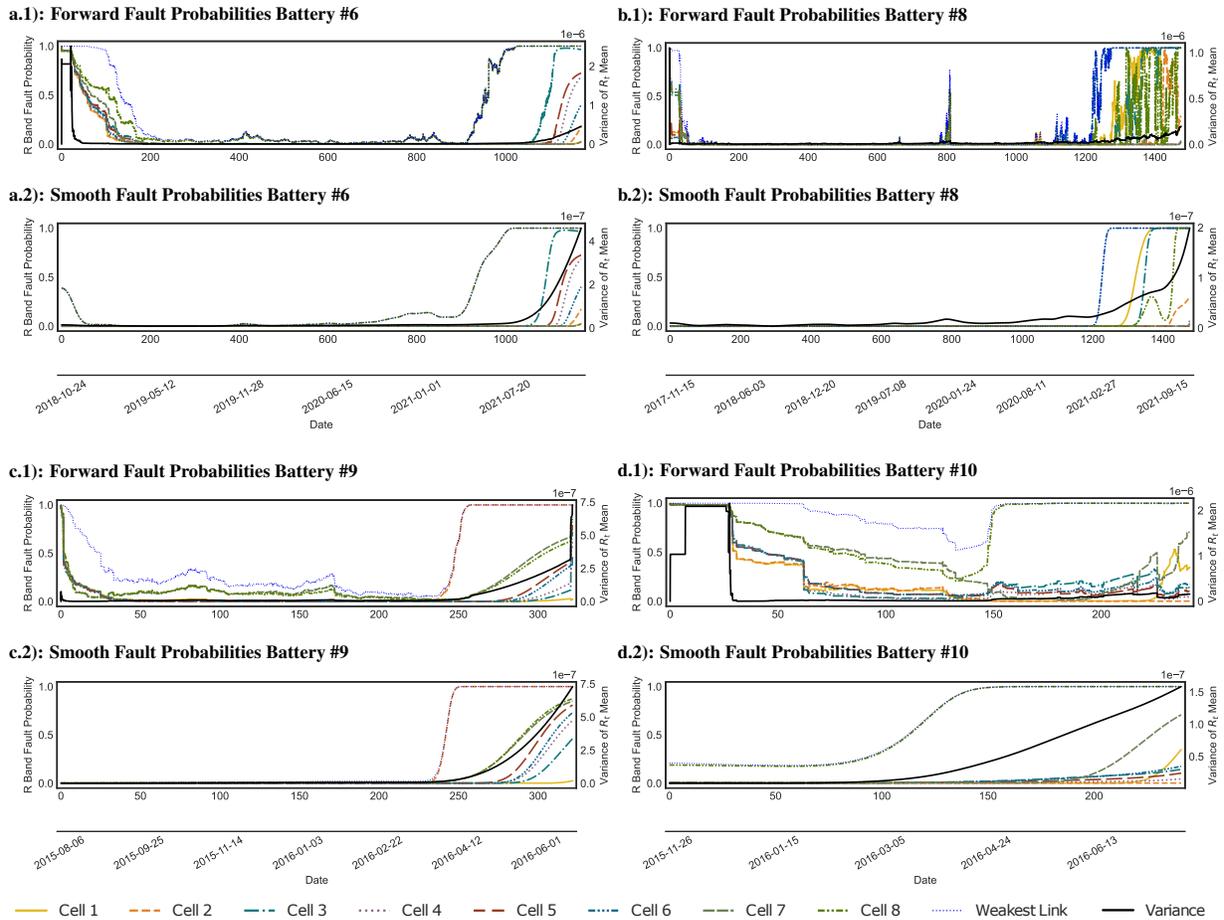
\begin{figure}[hbt] 
\centering
  \begin{subfigure}[b]{0.49\linewidth}
    \centering
    \begin{tikzpicture}
        \node at (0,0) (image1) {\includegraphics[trim={0 3cm 0 0cm}, clip, width=.99\linewidth]{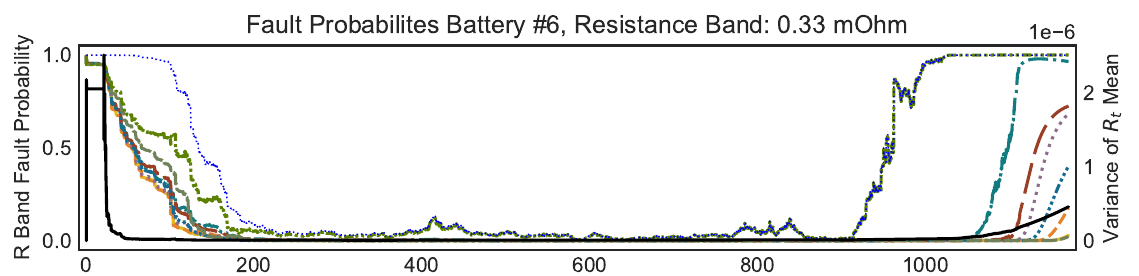}};
        \node [anchor=north west, xshift=-120pt, yshift=35pt] {\scriptsize\textbf{a.1): Forward Fault Probabilities Battery \#6}};
        \fill [white] (-2.5,0.68) rectangle (2.5,0.87);
    \end{tikzpicture} 
    \label{fig:fb:a1} 
    \vspace{-2ex}
  \end{subfigure}%
  \begin{subfigure}[b]{0.49\linewidth}
    \centering
    \begin{tikzpicture}
        \node at (0,0) (image1) {\includegraphics[trim={0 3cm 0 0cm}, clip, width=.99\linewidth]{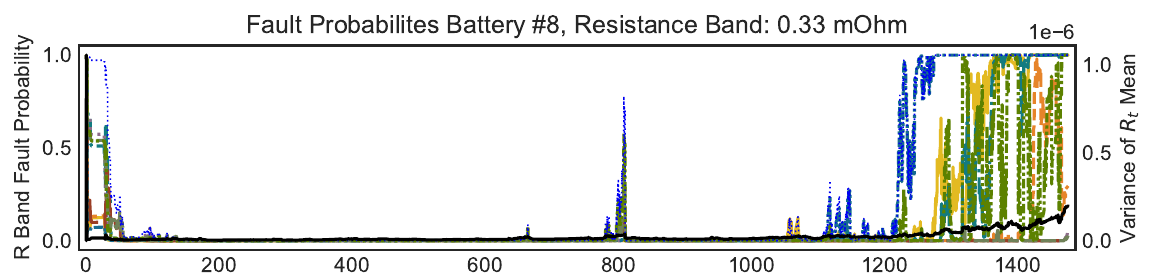}};
        \node [anchor=north west, xshift=-120pt, yshift=35pt] {\scriptsize\textbf{b.1): Forward Fault Probabilities Battery \#8}};
        \fill [white] (-2.5,0.68) rectangle (2.5,0.87);
    \end{tikzpicture} 
    \label{fig:fb:b1} 
    \vspace{-2ex}
  \end{subfigure}
  \begin{subfigure}[b]{0.49\linewidth}
    \centering
    \begin{tikzpicture}
        \node at (0,0) (image1) {\includegraphics[width=.99\linewidth]{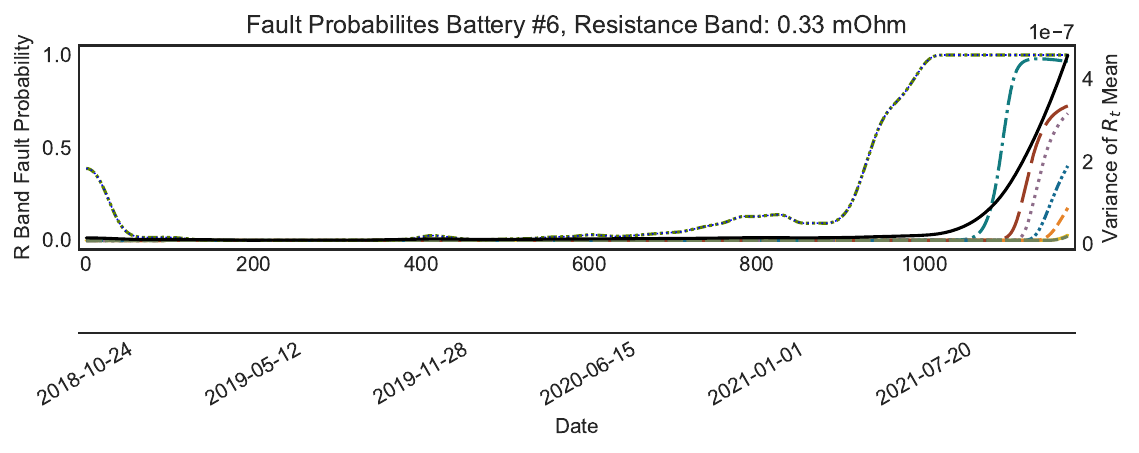}};
        \node [anchor=north west, xshift=-120pt, yshift=53pt] {\scriptsize\textbf{a.2): Smooth Fault Probabilities Battery \#6}};
        \fill [white] (-2.5,1.3) rectangle (2.5,1.5);
    \end{tikzpicture} 
    \label{fig:fb:a2} 
    \vspace{-2ex}
  \end{subfigure}%
  \begin{subfigure}[b]{0.49\linewidth}
    \centering
    \begin{tikzpicture}
        \node at (0,0) (image1) {\includegraphics[width=.99\linewidth]{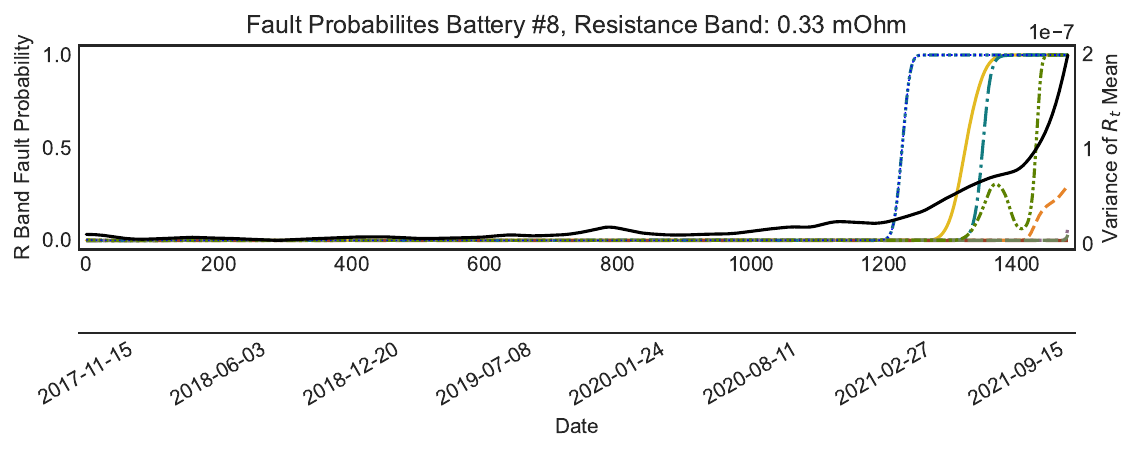}};
        \node [anchor=north west, xshift=-120pt, yshift=53pt] {\scriptsize\textbf{b.2): Smooth Fault Probabilities Battery \#8}};
        \fill [white] (-2.5,1.3) rectangle (2.5,1.5);
    \end{tikzpicture} 
    \label{fig:fb:b2} 
    \vspace{-2ex}
  \end{subfigure}
  \begin{subfigure}[b]{0.49\linewidth}
    \centering
    \begin{tikzpicture}
        \node at (0,0) (image1) {\includegraphics[trim={0 3cm 0 0cm}, clip, width=.99\linewidth]{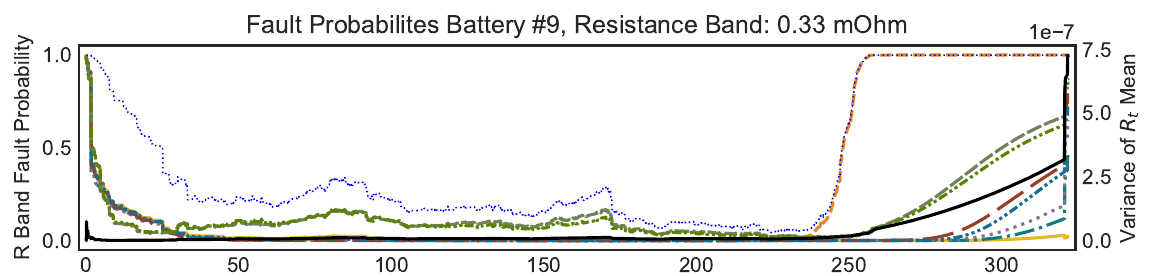}};
        \node [anchor=north west, xshift=-120pt, yshift=35pt] {\scriptsize\textbf{c.1): Forward Fault Probabilities Battery \#9}};
        \fill [white] (-2.5,0.68) rectangle (2.5,0.87);
    \end{tikzpicture} 
    \label{fig:fb:c1} 
    \vspace{-2ex}
  \end{subfigure}%
  \begin{subfigure}[b]{0.49\linewidth}
    \centering
    \begin{tikzpicture}
        \node at (0,0) (image1) {\includegraphics[trim={0 3cm 0 0cm}, clip, width=.99\linewidth]{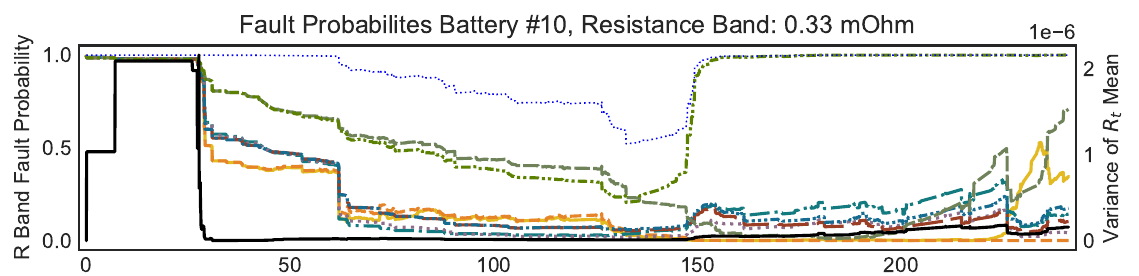}};
        \node [anchor=north west, xshift=-120pt, yshift=35pt] {\scriptsize\textbf{d.1): Forward Fault Probabilities Battery \#10}};
        \fill [white] (-2.5,0.68) rectangle (2.5,0.87);
    \end{tikzpicture} 
    \label{fig:fb:d1} 
    \vspace{-2ex}
  \end{subfigure}
  \begin{subfigure}[b]{0.49\linewidth}
    \centering
    \begin{tikzpicture}
        \node at (0,0) (image1) {\includegraphics[width=.99\linewidth]{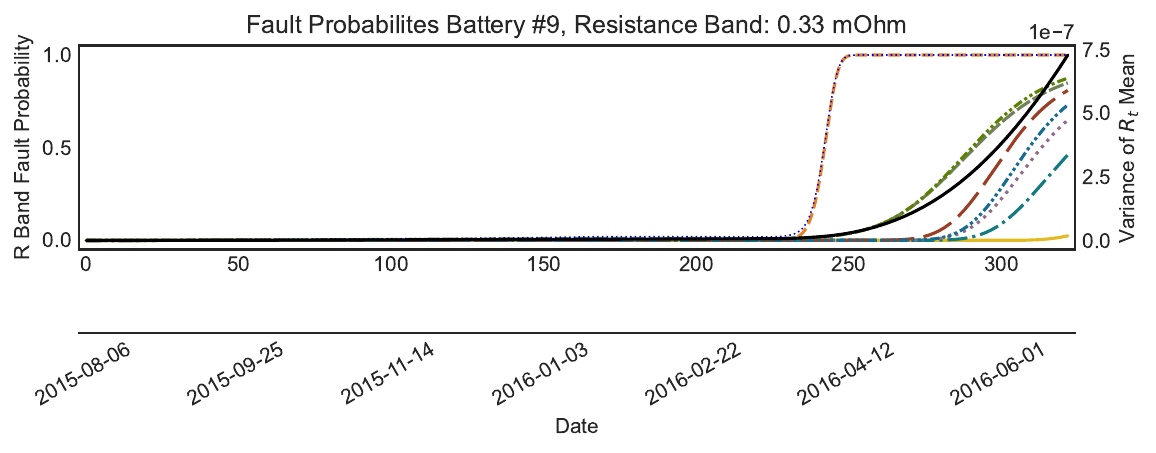}};
        \node [anchor=north west, xshift=-120pt, yshift=53pt] {\scriptsize\textbf{c.2): Smooth Fault Probabilities Battery \#9}};
        \fill [white] (-2.5,1.3) rectangle (2.5,1.5);
    \end{tikzpicture} 
    \label{fig:fb:c2} 
    \vspace{-4ex}
  \end{subfigure}%
  \begin{subfigure}[b]{0.49\linewidth}
    \centering
    \begin{tikzpicture}
        \node at (0,0) (image1) {\includegraphics[width=.99\linewidth]{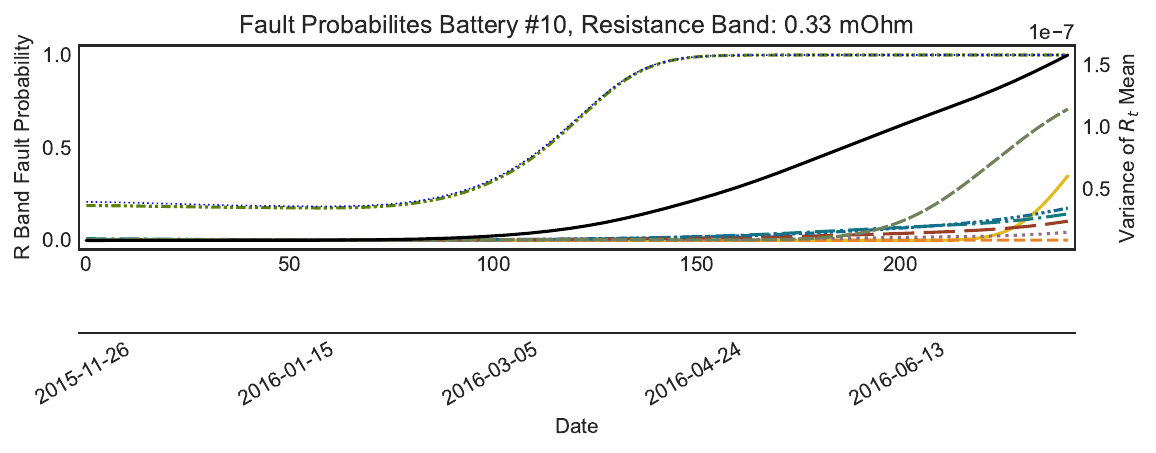}};
        \node [anchor=north west, xshift=-120pt, yshift=53pt] {\scriptsize\textbf{d.2): Smooth Fault Probabilities Battery \#10}};
        \fill [white] (-2.5,1.3) rectangle (2.5,1.5);
    \end{tikzpicture} 
    \label{fig:fb:d2} 
    \vspace{-4ex}
  \end{subfigure}
  \begin{subfigure}[b]{0.99\linewidth}
    \centering
     \begin{tikzpicture}
    \node[anchor=south west] at (0,0) (image1) {\includegraphics[width=.99\linewidth]{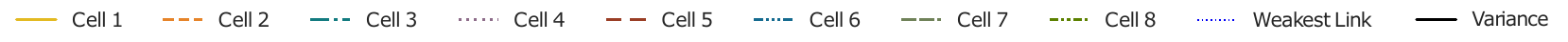} };
    \end{tikzpicture} 
  \end{subfigure}
  \caption{Fault probabilities for each cell of the four systems with the highest equivalent full cycle count in color. Variance of the \gls{gp} resistance mean in black. a) System 6, b) System 8, c) System 10, d) System 12.}
  \label{fig:fault_probs} 
\end{figure}

The forward fault probabilities (Figs.\,\ref{fig:fault_probs}a.1,b.1,c.1,d.1) fluctuate more than the smooth fault probabilities (Figs.\,\ref{fig:fault_probs}a.2,b.2,c.2,d.2), which is expected from the fact that they only depend on past data. In addition, the forward fault probabilities need a certain amount of data to settle in, here in the order of 30-150 days, depending on the usage of the system. 
The fault probabilities for Systems 6, 8, and 9 show similar patterns with very low fault probabilities for most of their usage (Fig.\,\ref{fig:fault_probs}). Generally, the forward and smooth fault probabilities agree well and detect the onset of high fault probabilities close to 1 at the same point in time.

\textbf{\textit{System 6:}} The fault probability of cell 8 increases after 600 days and crosses 0.5 shortly before 1000 days. The remaining cells also start deviating more from one another, as seen by their increasing resistance fault probabilities. Interestingly, cell 8 showed already a small fault probability at the beginning of life.

\textit{\textbf{System 8:}} The mean resistances associated with system 8 show seasonal variations (Fig.\,\ref{fig:all_batts_analysis}h), in line with a wide temperature distribution (Fig.\,\ref{fig:joy_us}d) where most readings are in a range between 10 and $45^\circ \mathrm{C}$, a lot wider than the distributions of most other systems. Around day 1000, the resistance fault probabilities of cells 6, 7, and 8 start to increase sharply. Figure \ref{fig:all_batts_analysis} shows a knee-shaped resistance with a fairly similar knee point, suggesting that most cells reached the region of accelerated degradation at about the same time.

\textbf{\textit{System 9:}} System 10 shows a similar pattern to system 6. Here, cell 2 displays a knee in the mean resistance at day 220 in the analyzed section (Fig.\,\ref{fig:all_batts_analysis}).

\textbf{\textit{System 10:}} Cells 7 and 8 have a significant fault probability already from the start, suggesting that these cells had an issue early on, in line with the end of use after 211 equivalent full cycles (Tab.\,\ref{tab:batteries_interpret}).

The fault probabilities allow us to analyze the aging in detail and unlock online monitoring.

\section*{Conclusion}
\label{sec:conclusion}
Health monitoring, fault analysis, and detection are critical for the safe and sustainable operation of battery systems. To analyze and detect faults, we developed a recursive and exact Gaussian process electrical circuit modeling pipeline to estimate the time-dependent equivalent circuit internal resistance as a surrogate for \gls{soh} from lithium-ion battery pack field data with measurement noise and unknown \gls{ocv} curve. Furthermore, we developed resistance fault probabilities using the individual cell resistance of the cells in a pack, suitable for early online detection of resistive knee points and resistance faults.

The methods are motivated and tested on a large field data set comprising 29 battery systems and more than 131 million data rows. The results show that often, a single cell with abnormal performance can cause the end of a system's use and that such faults can be detected with the proposed Gaussian process electrical circuit modeling approach. Such an occurrence can be caused after heavy use due to degradation or due to other issues. The results support that if a cell has a higher resistance early on, this can already be an indicator that this cell will age faster than the remaining cells. Physically, this could occur by enhanced resistive heating of the most degraded cell, thereby accelerating its degradation~\cite{gogoana2014internal}, which has been observed in the case of Li-ion batteries connected in series~\cite{chiu2014cycle}. 
Furthermore, the results show the promise of online cell monitoring and suggest that, for larger battery packs, advanced pack layouts that allow the permanent bypass of a more strongly degraded cell can be advantageous in prolonging the life of a system.

To conclude, this analysis furthers the understanding of how batteries degrade and fail in the field and demonstrates the potential of online monitoring. 
We publish the associated data and Python software \textit{BattGP} upon completion of the review of this article, making crucially important data and tools for analyzing and modeling battery field data available to the community.

\vspace{8ex}
\section*{Methods}
\label{sec:methods}
\subsection*{Data Selection} The data selection criteria (Tab.\,\ref{tab:data_selection}) are applied to each cell individually.
\begin{table}[H]
    \centering
    \caption{Data Selection Criteria}
    \label{tab:data_selection}
    \vspace{1mm}
    \begin{tabular}{l|rcl}
    Charging/Discharging & \multicolumn{3}{l}{Discharge only} \\
    \hline
    Temperature ($^\circ$C) & $10$ & \hspace{-4mm} $< x < $ &  \hspace{-4mm} $100$ \\
    \hline
    Current (A) & $-200$ & \hspace{-4mm} $< x < $ &  \hspace{-4mm} $-5$\\
    \hline
    SOC ($\%$) & $40$ & \hspace{-4mm} $< x <$ &  \hspace{-4mm} $94$ \\
    \end{tabular}
\end{table}

\subsection*{GPs}
\Glspl{gp} are a flexible modeling framework excelling in the case of limited data by making a point estimate and modeling the covariance associated with the prediction. \Glspl{gp} are nonparametric probabilistic models fully defined by a mean function $\mu(x)$ and a covariance function $k(x, x')$ where $x,x' \in \mathcal{X}^D$.
A \gls{gp} is usually written as
\begin{equation}
    f(x) \sim \mathrm{GP}(\mu(x), k(x, x')),
\end{equation}
making it explicit that \glspl{gp} describe a distribution of functions. The \gls{gp} posterior predictions are normally distributed, i.e., any marginal distribution of the \gls{gp} is Gaussian. Furthermore, all joint distributions associated with a finite number of elements of the index set are multivariate normal distributions (e.g., \cite{williams2006gaussian} for further information).
Assuming that we have $n_\mathrm{o}$ noisy observations $(x_{\mathrm{o},i}, y_{\mathrm{o},i})$ with $y_{\mathrm{o},i}=f(x_\mathrm{o,i})+\epsilon_i$, where $\epsilon_i$ is Gaussian noise with variance $\sigma_\mathrm{n}^2$,
the predictive \gls{gp} equations are
\begin{align}
    \mu_{|\mathrm{o}}(X_*) &= K(X_*, X_\mathrm{o})[K(X_\mathrm{o},X_\mathrm{o}) + \sigma_\mathrm{n}^2 I]^{-1}y_\mathrm{o} \label{eq:gp_p_mean}\\
    \Sigma_{|\mathrm{o}}(X_*) &= K(X_*, X_*) - K(X_*, X_\mathrm{o})[K(X_\mathrm{o},X_\mathrm{o}) + \sigma_n^2I]^{-1}K(X_\mathrm{o}, X_*)
    \label{eq:gp_p_var},
\end{align}
where $X_*=\begin{bmatrix} x_{*,1} & \cdots & x_{*,n_*} \end{bmatrix}$ denotes the $n_*$ test locations, $X_\mathrm{o}=\begin{bmatrix} x_{\mathrm{o},1} & \cdots & x_{\mathrm{o},n_\mathrm{o}} \end{bmatrix}$ denotes the training locations with responses $y_\mathrm{o}^\mathrm{T} = \begin{bmatrix} y_{\mathrm{o},1} & \cdots & y_{\mathrm{o},n_\mathrm{o}} \end{bmatrix}$, and $K(X_1, X_2)$ denotes the covariance matrix that is constructed by applying the kernel function to all pairs of column vectors from $X_1$ and $X_2$ (for a full derivation, see \cite{williams2006gaussian}).

The training of a \gls{gp} refers to the choice of kernel function and the optimization of associated hyperparameters, usually based on optimizing the marginal likelihood of the training data. Subsequently, the posterior distribution can be calculated for points of interest by inference, using (\ref{eq:gp_p_mean},~\ref{eq:gp_p_var}).

\textit{SE Kernel}: The \gls{se} kernel is a smooth, infinitely many times differentiable kernel which only depends on the distance of data points and is given for one-dimensional $x$ by
\begin{equation}
    k_\text{SE}(x, x') = \sigma^2_\mathrm{SE} \exp\!\left(\frac{-|x-x'|^2}{2l^2}\right).
\end{equation}
We use three input dimensions -- current, \gls{soc}, and temperature -- and combine three one-dimensional \gls{rbf} kernels to the third-order additive kernel,
\begin{equation}
    k_\text{SE,3ARD}(x, x') = \prod_{d=1}^3 \sigma^2_\text{SE,d} \exp\!\left(\frac{-|x_d-x_d'|^2}{2l_d^2}\right) = \sigma^2_\text{SE,3} \exp\!\left(\sum_{d=1}^3\frac{-|x_d-x_d'|^2}{2l_d^2}\right),
\end{equation}
which is known as the \gls{se}-ARD kernel (ARD denotes ``automatic relevance detection'' because each length scale represents the importance of its associated direction; for more information, see \cite{duvenaud2014automatic}). There are four hyperparameters associated with the three-dimensional \gls{se}-ARD kernel: the output scale $\sigma_\text{SE, 3}^2$, the length scale associated with the current dimension $l_1 = l_\text{I}$, the length scale associated with the \gls{soc} dimension $l_2 = l_\text{SOC}$, and the length scale associated with the temperature dimension $l_3 = l_\text{T}$.

\textit{Wiener Velocity Kernel}: The \gls{wv} model corresponds to the integrated Wiener process \cite{Solin}. The \gls{wv} covariance function or kernel is defined by
\begin{equation}
    k_{\mathrm{WV}}(t, t^{\prime})=\sigma_\text{WV}^2\!\left(\frac{\min ^3\{t, t^{\prime}\}}{3}+\left|t-t^{\prime}\right| \frac{\min ^2\{t, t^{\prime}\}}{2}\right).
    \label{eq:wiener_kernel}
\end{equation}

The \gls{wv} kernel has one hyperparameter: the \gls{wv} output scale, $\sigma_\text{WV}^2$.

Including the noise variance $\sigma_\mathrm{n}^2$ there are six hyperparameters, which define the characteristics of the \gls{gp} used in this article.

\newcommand{\Kmm}{K_\mathrm{mm}}
\newcommand{\Kmb}{K_\mathrm{mb}}
\newcommand{\Kbb}{K_\mathrm{bb}}
\newcommand{\Kqq}{K_\mathrm{qq}}
\newcommand{\Kqb}{K_\mathrm{qb}}

\newcommand{\kt}{k_\mathrm{t}}
\newcommand{\ks}{k_\mathrm{s}}

\newcommand{\x}{x}
\newcommand{\xt}{t}
\newcommand{\xs}{x_\mathrm{s}}

\newcommand{\Xm}{X_{\mathrm{m},k}}
\newcommand{\Xb}{X_\mathrm{b}}
\newcommand{\Xq}{X_\mathrm{q}}

\newcommand{\ym}{y_{\mathrm{m},k}}

\newcommand{\nm}{n_{\mathrm{m},k}}
\newcommand{\nb}{n_\mathrm{b}}
\newcommand{\nq}{n_\mathrm{q}}

\newcommand{\Ht}{H_\mathrm{t}}
\newcommand{\Hs}{H_\mathrm{s}}

\newcommand{\zt}{z_\mathrm{t}}
\newcommand{\zs}{z_\mathrm{s}}

\newcommand{\Ts}{T_\mathrm{s}}

\newcommand{\transp}{\mathrm{T}}

\paragraph{Spatiotemporal GPs}

A large number of data points are recorded during the operation of a battery. Ideally, all samples that fulfill the data selection criteria are used, but a classical exact \gls{gp} prevents this due to the unfavorable scaling of compute and memory usage.
To address this issue, we follow \cite{huber2014recursive, Solin} and leverage the fact that we can interpret the \gls{gp} as a spatiotemporal \gls{gp}. For lead-acid batteries, a spatiotemporal \glspl{gp} approach has been shown by \cite{aitioSolar2021}.

A temporal \gls{gp}, i.e., a \gls{gp} that depends on only one variable, can be written as a Kalman filter \cite{sarkka_spatiotemp, Solin}, which makes it scale linearly with the number of data points, with the restriction that the data points must be given in a sorted manner. Here, the variable is actually time; therefore, this does not impose further restrictions.

A spatiotemporal \gls{gp} \cite{Solin} is a \gls{gp} that depends on multiple variables, of which one can be regarded as the time variable, the others being the spatial variables,
\begin{align*}
  \x = \begin{bmatrix} \xt \\ \xs \end{bmatrix}\!.
\end{align*}
This concept leads to an infinite-dimensional Kalman filter that propagates the mean and variance functions over the time points.

A further important simplification arises as our model is separable, i.e., we can write the covariance function as a product
\begin{align*}
  k(x, x) = \kt(\xt, \xt) \ks(\xs, \xs).
\end{align*}
When using only a finite number of spatial vectors, this eventually leads to a classical, finite-dimensional Kalman filter \cite{Solin}.

We cannot restrict the actual measured data locations to a finite set. Therefore, we combine the spatiotemporal approach with the recursive approach from Huber \cite{huber2014recursive}.
In this approach the spatial part of the \gls{gp} is not represented exactly, but it is represented by $\nb$ predefined basis vectors $x_{\mathrm{s,b},i}$ that are collected into
\begin{align*}
  \Xb =
    \begin{bmatrix}
      x_{\mathrm{s,b},1} & x_{\mathrm{s,b},2} & \cdots & x_{\mathrm{s,b},n_\mathrm{b}}
    \end{bmatrix}\!.
\end{align*}

Combining these two approaches leads to the following procedure, which is similar but not identical to the method used in \cite{aitioSolar2021}. The difference is mentioned below.

The state $z$ of the Kalman filter,
\begin{align*}
  z = \begin{bmatrix} \zt \\ \zs \end{bmatrix}\!,
\end{align*}
comprises two parts. $\zt$ corresponds to the representation of the temporal kernel.
In the case of the Wiener velocity kernel, it contains two scalar values; the first represents the mean value, and the second is the time derivative of the mean value \cite{Solin}.
$\zs$ corresponds to the spatial kernel, and for the Huber approach \cite{huber2014recursive}, these are the mean values at the specified basis vectors $\Xb$.

\paragraph{Evaluation}
Given the state $z_{k|k}$ and its covariance matrix $P_{k|k}$ at the time $t_k$ and built with all information available until $t_k$, the mean and covariance at this time point can be calculated by
\begin{align}
  \mu_{|k}(\Xq, t_k) & = H z_{k|k} \label{eq:stgp:mu_xq}\\
  \Sigma_{|k}(\Xq, t_k) & = \Kqq + H P_{k|k} H^\transp - \Hs \Kbb \Hs^\transp. \label{eq:stgp:sigma_xq}
\end{align}
The mean and the covariance of the \gls{gp} at $\nq$ spatial vectors $x_{\mathrm{s,q},i}$ are collected into $\Xq$ similar to $\Xb$. Furthermore, we use the abbreviations
\begin{align*}
  \Kbb & \in \mathbb{R}^{\nb \times \nb}, \quad (\Kbb)_{i,j} = \ks(x_{\mathrm{s,b},i}, x_{\mathrm{s,b},j}) ,\\
  \Kqq & \in \mathbb{R}^{\nq \times \nq}, \quad (\Kqq)_{i,j} = \ks(x_{\mathrm{s,q},i}, x_{\mathrm{s,q},j}),
\end{align*}
for the covariance at the basis vectors and the covariance at the queried locations, respectively.

The measurement matrix
\begin{align}
  H =
    \begin{bmatrix}
      \Ht & \Hs
    \end{bmatrix}
  \label{eq:stgp:h}
\end{align}
is partitioned according to $z$. The temporal part
\begin{align}
  \Ht & =
    \begin{bmatrix}
      1 & 0 \\
      \vdots & \vdots \\
      1 & 0
    \end{bmatrix}\!,
  \label{eq:stgp:ht}
\end{align}
reflects that for the Wiener velocity kernel with the chosen representation, the first associated state corresponds to the mean. The spatial part is
\begin{align}
  \Hs & =
    \Kqb \Kbb^{-1},
  \label{eq:stgp:hs}
\end{align}
with
\begin{align*}
  \Kqb & \in \mathbb{R}^{\nq \times \nb}, \quad (\Kqb)_{i,j} = \ks(x_{\mathrm{s,q},i}, x_{\mathrm{s,b},j})
\end{align*}
given by \cite{huber2014recursive}.
Here, our approach differs from the approach described in \cite{aitioSolar2021}.
In \eqref{eq:stgp:hs}, only the inverse of $\Kbb$ is needed, which is known a priori and can be calculated offline. In \cite{aitioSolar2021}, the corresponding term includes an additional part that depends on the current state; thus, the inverse must be calculated in each step anew.

Equation \eqref{eq:stgp:mu_xq} shows that the assumption of a separable kernel means that the mean values given by the temporal kernel and the spatial kernel are added, which is in accordance with the modeling of the two resistances in series (see Fig.\,\ref{fig:gp-ecm-model}).

\paragraph{Prediction step}
The prediction step of the Kalman filter is performed identically to \cite{aitioSolar2021},
\begin{align*}
  z_{k|k-1} & = A(\Ts) z_{k-1|k-1} \\
  P_{k|k-1} & = A(\Ts) P_{k-1|k-1} A^\transp(\Ts) + Q(\Ts)
\end{align*}
with
\begin{align*}
  P(0) & =
    \begin{bmatrix}
      P_\mathrm{t}(0) & 0 \\ 0 & P_\mathrm{s}(0)
    \end{bmatrix},
  &
  A(\Ts) & =
    \begin{bmatrix}
      A_\mathrm{t}(\Ts) & 0 \\ 0 & I
    \end{bmatrix},
  &
  Q(\Ts) & =
    \begin{bmatrix}
      Q_\mathrm{t}(\Ts) & 0 \\ 0 & 0
    \end{bmatrix},
\end{align*}
where $\Ts = t_k - t_{k-1}$ is the length of the time step that may be different for each prediction step.
The structure of these matrices shows the decoupling of the two parts, i.e., the spatial state is not influenced by the time step.
However, due to the correction step, the covariance matrix will not retain its initial block diagonal structure.

For the Wiener velocity kernel \eqref{eq:wiener_kernel}, which is used as temporal kernel, the corresponding matrices are
\begin{align*}
  P_\mathrm{t}(0) & =
    \begin{bmatrix}
      0 & 0 \\ 0 & 0
    \end{bmatrix},
  &
  A_\mathrm{t}(\Ts) & =
    \begin{bmatrix}
      1 & \Ts \\ 0 & 1
    \end{bmatrix},
  &
  Q_\mathrm{t}(\Ts) & =
    \sigma_\mathrm{WV}^2 
    \begin{bmatrix}
      \Ts^3/3 & \Ts^2/2 \\[2mm] \Ts^2/2 & \Ts
    \end{bmatrix}.
\end{align*}
The initial value for the spatial covariance matrix is $P_\mathrm{s}(0) = \Kbb$. \cite{Solin, aitioSolar2021}

\paragraph{Correction step}
When new measurements $(\Xm, \ym)$ arrive, the Kalman filter performs a correction step
\begin{align*}
  z_{k|k} & = z_{k|k-1} + K_k \left( \ym - H_k z_{k|k-1}\right) \\
  P_{k|k} & = P_{k|k-1} - K_k H_k P_{k|k-1}.
\end{align*}
$\Xm$ is a matrix that contains the locations of the $\nm$ measurements, $\ym$ is a vector giving the measured output for each location in $\Xm$, and $H_k$ is the measurement matrix build as in \eqref{eq:stgp:h} with $\Xq = \Xm$. The Kalman gain $K_k$ is calculated by 
\begin{align}
  K_k & = P_{k|k-1} H_k^\transp \left( \Sigma_{|k}(\Xm) + \sigma_\mathrm{n}^2 I \right)^{-1}\;,
  \label{eq:stgp:Kk}
\end{align}
where $\Sigma_{|k}(\Xm)$ is given by \eqref{eq:stgp:sigma_xq} with $\Xq = \Xm$.
This step differs from \cite{aitioSolar2021} by the different evaluation mentioned above and another measurement model.

\emph{Remark 1:}
If $\nm = 0$, i.e., no measurement data was acquired within the last time step, the correction step can be skipped, and the algorithm proceeds with the next prediction step.
Alternatively, the algorithm could wait until new data is available and then perform a larger prediction step.

\emph{Remark 2:}
In a standard Kalman filter setup, the correction step can be performed iteratively for each group of measurements that is not correlated with measurements outside this group.
Consequently, for the case here, where we model the measurement noise with the covariance matrix $\sigma_\mathrm{n}^2 I$, i.e., all measurements are uncorrelated, each measurement could be processed individually, thus rendering the matrix inverse in \eqref{eq:stgp:Kk} a scalar inversion.
However, due to some simplifications in the approach of \cite{huber2014recursive} that led to the filter equations, processing each measurement individually or multiple measurements together in a minibatch is 
not analytically equivalent for this application. Based on case studies, it seems to be advantageous to add measurements in larger groups.

\paragraph{Rauch-Tung-Striebel smoother}

The Rauch-Tung-Striebel smoother allows calculating the state and covariance and thus the estimates $\mu_{|n}(\Xq,t_k)$ and $\Sigma_{|n}(\Xq,t_k)$ at all time points $t_k$ under consideration of the complete measured data of the $n$ time points.
This smoother starts with the variables $z_{n|n}$ and $P_{n|n}$ of the last Kalman filter correction step and iterates backward using \cite{Sarkka2019_applied_sde}
\begin{align}
  z_{k|n} & = z_{k|k} + G_k ( z_{k+1|n} - z_{k+1|k} ) \label{eq:stgp:smoother_z} \\
  P_{k|n} & = P_{k|k} + G_k ( P_{k+1|n} - P_{k+1|k}) G_k^\transp \label{eq:stgp:smoother_P}
\end{align}
for $k=n-1,\ n-2,\ \ldots$ with
\begin{align}
  G_k = P_{k|k} A(t_{k+1} - t_k) P_{k+1|k}^{-1}\;.
  \label{eq:stgp:smoother_G}
\end{align}
The output values $\mu_{|n}(\Xq,t_k)$ and their variances $\Sigma_{|n}(\Xq,t_k)$ for each time point $t_k$ can be calculated by \eqref{eq:stgp:mu_xq} and \eqref{eq:stgp:sigma_xq} by replacing $z_{k|k}$ with $z_{k|n}$ and $P_{k|k}$ with $P_{k|n}$.

For evaluating equations \eqref{eq:stgp:smoother_z}, \eqref{eq:stgp:smoother_P} and \eqref{eq:stgp:smoother_G}, the results $z_{k|k}$ and $P_{k|k}$ of the Kalman filter are used. Thus, these values must be stored while performing the forward filtering pass.
$z_{k+1|k}$ and $P_{k+1|k}$ could either be stored also or calculated anew from $z_{k|k}$ and $P_{k|k}$ by reevaluating the prediction step.

\paragraph{Implementation for the presented results}

The batteries are mostly sampled with 5\,s, which, with the very small magnitude of sensible parameters for the output scale of the Wiener velocity kernel, leads to numerical issues in performing the prediction steps.

Therefore, we define a sampling time for the updates (1~hour in all shown results) and process all data captured within this interval at the same update time point.
If no data was recorded within this hour, we only perform a new prediction step without the update to set an evaluation point for the later backward smoothing pass.

\paragraph{Selection of Basis Vectors}
The selection of basis vectors can be done in different ways. Similarly to \cite{aitioSolar2021}, we found that using k-means to select the basis vectors is effective, i.e., the results and confidence intervals of the spatiotemporal \gls{gp} with only 27 basis vectors selected by k-means as well as the reference operating point, resulting in 28 basis vectors (SI Fig.\,B.1) resemble fairly closely the results of the full \gls{gp} with up to 40k data points. However, in an online setting, there will be no data available at the beginning. Therefore, as an alternative, we propose to place vectors linearly within the range of the data selection criteria and, furthermore, set a constraint based on the length scale of the associated measurement (SI Sec.\,B).

\subsection*{Hyperparameter Tuning}
All battery systems contain the same prismatic cell type with the same specifications. Therefore, the aim is to find a single hyperparameter set that can be used for all cells. %
We optimized the hyperparameters on the four longest-lived systems (6, 8, 9, and 10) by optimizing the marginal likelihood using 16k data points. The mode of each hyperparameter was chosen as a robust estimate, assuming that the parameters are mutually independent.
We do not formulate an optimization problem for our spatiotemporal \gls{gp} but use the hyperparameters found by the exact \gls{gp}. The results of the exact \gls{gp} using 40k data points (Fig. \ref{fig:all_batts_analysis}) and the spatiotemporal \gls{gp} using all selected data (SI Figs.\,B.1,B.2), gives confidence that this approach is reasonable.

\subsection*{Software and Hardware Details}
The full GP is solved on an NVIDIA A100 GPU with 80\,GB RAM. We use the Python GPytorch framework, allowing the solving of an exact \gls{gp} with 40k data points selected by their index within approximately a few minutes per battery system. The recursive \gls{gp} is solved on a MacBook Pro with a 10-core M1 chip, taking approximately 2 min for a single cell of system \#8 with over 1 million data points and less than 1 hour to process all data from the 21 analyzed systems.

\section*{Data and Code Availability}
\label{sec:data_code_avail}
We publish the associated data and Python software \textit{BattGP} upon completion of the review of this article. %

\section*{Author Contributions}
Joachim Schaeffer: Conceptualization, Methodology, Software, Validation, Formal Analysis, Investigation, Data Curation, Writing -- original draft, Writing -- review \& editing, Visualization;
Eric Lenz: Methodology, Software, Formal Analysis, Writing -- review \& editing, Visualization;
Duncan Gulla: Methodology, Software, Investigation, Data Curation, Writing -- review \& editing;
Martin Z. Bazant: Writing -- review \& editing;
Richard D. Braatz: Writing -- review \& editing, Supervision;
Rolf Findeisen: Resources, Writing -- review \& editing, Supervision, Funding Acquisition

\section*{Acknowledgments}
We would like to thank numerous people for their outstanding support throughout this project; however, we can not name them due to data privacy requirements from the data provider. In particular, we are truly grateful for the outstanding support from our main contacts from the data provider; without you, this work would not have been possible.

\section*{Funding}
The work was financed through base funding from the Technical University of Darmstadt.

\section*{Competing Interests}
None.

\section*{Supplementary Information}
The supplementary information is only available in version 1 on arXiv due to its large size; version 2 (this version) is otherwise identical to version 1.

\newpage

\printbibliography

\end{document}